\documentclass[twoside,11pt]{article}

\usepackage{jmlr2e}
\usepackage{macros}

\ShortHeadings{Memory Augmented Neural Networks with Wormhole Connections}{Gulcehre, Chandar, and Bengio}
\firstpageno{1}

\begin{document}

\title{Memory Augmented Neural Networks with Wormhole Connections}

\author{\name Caglar Gulcehre \email gulcehrc@iro.umontreal.ca\\
\addr Montreal Institute for Learning Algorithms\\ 
       Universite de Montreal\\
       Montreal, Canada
       \AND
       \name Sarath Chandar \email apsarathchandar@gmail.com \\
       \addr Montreal Institute for Learning Algorithms\\ 
       Universite de Montreal\\
       Montreal, Canada
       \AND
       \name Yoshua Bengio \email yoshua.bengio@umontreal.ca\\
       \addr Montreal Institute for Learning Algorithms\\ 
       Universite de Montreal\\
       Montreal, Canada}

\editor{Leslie Pack Kaelbling}

\maketitle

\begin{abstract}
Recent empirical results on long-term dependency tasks have shown that neural networks augmented with an external memory can learn the long-term dependency tasks more easily and achieve better generalization than vanilla recurrent neural networks (RNN). We suggest that memory augmented neural networks can reduce the effects of vanishing gradients by creating shortcut (or wormhole) connections. Based on this observation, we propose a novel memory augmented neural network model called TARDIS (Temporal Automatic Relation Discovery in Sequences). The controller of TARDIS can store a selective set of embeddings of its own previous hidden states into an external memory and revisit them as and when needed. For TARDIS, memory acts as a storage for wormhole connections to the past to propagate the gradients more effectively and it helps to learn the temporal dependencies. The memory structure of TARDIS has similarities to both Neural Turing Machines (NTM) and Dynamic Neural Turing Machines (D-NTM), but both read and write operations of TARDIS are simpler and more efficient. We use discrete addressing for read/write operations which helps to substantially to reduce the vanishing gradient problem with very long sequences. Read and write operations in TARDIS are tied with a heuristic once the memory becomes full, and this makes the learning problem simpler when compared to NTM or D-NTM type of architectures. We provide a detailed analysis on the gradient propagation in general for MANNs. We evaluate our models on different long-term dependency tasks and report competitive results in all of them.
\end{abstract}

\section{Introduction}
\label{sec:intro}

Recurrent Neural Networks (RNNs) are neural network architectures that are designed to handle temporal dependencies in sequential prediction problems. However it is well known that RNNs suffer from the issue of vanishing gradients as the length of the sequence and the dependencies increases \citep{hochreiter1991untersuchungen,bengio1994learning}. Long Short Term Memory (LSTM) units \citep{lstm1997} were proposed as an alternative architecture which can handle long range dependencies better than a vanilla RNN. A simplified version of LSTM unit called Gated Recurrent Unit (GRU), proposed in \citep{cho2014learning}, has proven to be successful in a number of applications \citep{bahdanau2014neural,xu2015show,trischler2016natural,kaiser2015neural,serban2016building}. Even though LSTMs and GRUs attempt to solve the vanishing gradient problem, the memory in both architectures is stored in a single hidden vector as it is done in an RNN and hence accessing the information too far in the past can still be difficult. In other words, LSTM and GRU models have a limited ability to perform a search through its past memories when it needs to access a relevant information for making a prediction. Extending the capabilities of neural networks with a memory component has been explored in the literature on different applications with different architectures \citep{weston2014memory,graves2014neural,joulin2015inferring,grefenstette2015learning,sukhbaatarend,bordes2015large,chandar2016hierarchical,gulcehre2016dynamic,Graves_Nature2016,rae2016scaling}.

Memory augmented neural networks (MANN) such as neural Turing machines (NTM) \citep{graves2014neural,rae2016scaling}, dynamic NTM (D-NTM) \citep{gulcehre2016dynamic}, and Differentiable Neural Computers (DNC) \citep{Graves_Nature2016} use an external memory (usually a matrix) to store information and the MANN's controller can learn to both read from and write into the external memory. As we show here, it is in general possible to use particular MANNs to explicitly store the previous hidden states of an RNN in the memory and that will provide shortcut connections through time, called here {\it \textbf{wormhole connections}}, to look into the history of the states of the RNN controller. Learning to read and write into an external memory by using neural networks gives the model more freedom or flexibility to retrieve information from its past, forget or store new information into the memory. However, if the addressing mechanism for read and/or write operations are continuous (like in the NTM and continuous D-NTM), then the access may be too diffuse, especially early on during training. This can hurt especially the {\em writing} operation, since a diffused write operation will overwrite a large fraction of the memory at each step, yielding fast vanishing of the memories (and gradients). On the other hand, discrete addressing, as used in the discrete D-NTM, should be able to perform this search through the past, but prevents us from using straight backpropagation for learning how to choose the address.  

We investigate the flow of the gradients and how the wormhole connections introduced by the controller effects it. Our results show that the wormhole connections created by the controller of the MANN can significantly reduce the effects of the vanishing gradients by shortening the paths that the signal needs to travel between the dependencies. We also discuss how the MANNs can generalize to the sequences longer than the ones seen during the training.

In a discrete D-NTM, the controller must learn to read from and write into the external memory by itself and additionally, it should also learn the reader/writer synchronization. This can make the learning to be more challenging. In spite of this difficulty, \cite{gulcehre2016dynamic} reported that the discrete D-NTM can learn faster than the continuous D-NTM on some of the bAbI tasks. We provide a formal analysis of gradient flow in MANNs based on discrete addressing and justify this result. In this paper, we also propose a new MANN based on discrete addressing called TARDIS (Temporal Automatic Relation Discovery in Sequences). In TARDIS, memory access is based on tying the write and read heads of the model after memory is filled up. When the memory is not full, the write head store information in memory in the sequential order.

The main characteristics of TARDIS are as follows, TARDIS is a simple memory augmented neural network model which can represent long-term dependencies efficiently by using a external memory of small size. TARDIS represents the dependencies between the hidden states inside the memory. We show both theoretically and experimentally that TARDIS fixes to a large extent the problems related to long-term dependencies. Our model can also store sub-sequences or sequence chunks into the memory. As a consequence, the controller can learn to represent the high-level temporal abstractions as well. TARDIS performs well on several structured output prediction tasks as verified in our experiments. 

The idea of using external memory with attention can be justified with the concept of mental-time travel which humans do occasionally to solve daily tasks. In particular, in the cognitive science literature, the concept of chronesthesia is known to be a form of consciousness which allows human to think about time subjectively and perform mental time-travel~\citep{tulving2002chronesthesia}. TARDIS is inspired by this ability of humans which allows one to look up past memories and plan for the future using the episodic memory.

\section{TARDIS: A Memory Augmented Neural Network}
\label{sec:ext_memory}

Neural network architectures with an external memory represent the memory in a matrix form, such that at each time step $t$ the model can both read from and write to the external memory. The whole content of the external memory can be considered as a generalization of hidden state vector in a recurrent neural network. Instead of storing all the information into a single hidden state vector, our model can store them in a matrix which has a higher capacity and with more targeted ability to substantially change or use only a small subset of the memory at each time step. The neural Turing machine (NTM) \citep{graves2014neural} is such an example of a MANN, with both reading and writing into the memory.

\subsection{Model Outline}

In this subsection, we describe the basic structure of TARDIS~\footnote{Name of the model is inspired from the time-machine in a popular TV series Dr. Who.} (Temporal Automatic Relation Discovery In Sequences). TARDIS is a MANN which has an external memory matrix $\mM_t \in \mathbb{R}^{k \times q}$ where $k$ is the number of memory cells and $q$ is the dimensionality of each cell. The model has an RNN controller which can read and write from the external memory at every time step. To read from the memory, the controller generates the read weights $\vw^r_t \in \mathbb{R}^{k \times 1}$ and the reading operation is typically achieved by computing the dot product between the read weights $\vw^r_t$ and the memory $\mM_t$, resulting in the content vector $\vr_t \in \mathbb{R}^{q \times 1}$: 

\begin{equation} 
    \label{eqn:reading_linear}
    \vr_t= (\mM_t)^{\top}\vw^r_t,
\end{equation}
TARDIS uses discrete addressing and hence $\vw^r_t$ is a one-hot vector and the dot-product chooses one of the cells in the memory matrix \citep{rlntm,gulcehre2016dynamic}. The controller generates the write weights $\vw^w_t \in \mathbb{R}^{1 \times k}$, to write into the memory which is also a one hot vector, with discrete addressing. We will omit biases from our equations for the simplicity in the rest of the paper. Let $i$ be the index of the non-zero entry in the one-hot vector $\vw^w_t$, then the controller writes a linear projection of the current hidden state to the memory location $\mM_t[i]$:
\begin{equation}
\mM_t[i] = \mW_m \vh_t,
\label{eqn:linproj}
\end{equation}
where $\mW_m \in \R^{d_m \times d_h}$ is the projection matrix that projects the $d_h$ dimensional hidden state vector to a $d_m$ dimensional micro-state vector such that $d_h > d_m$.

At every time step, the hidden state $\vh_t$ of the controller is also conditioned on the content $\vr_t$ read from the memory. The wormhole connections are created by conditioning $\vh_t$ on $\vr_t$:
\begin{equation}
\vh_t = \phi(\vx_t, \vh_{t-1}, \vr_t).
\end{equation}

As each cell in the memory is a linear projection of one of the previous hidden states, the conditioning of the controller's hidden state with the content read from the memory can be interpreted as a way of creating short-cut connections across time (from the time $t'$ that $\vh_{t'}$ was written to the time $t$ when it was read through $\vr_t$) which can help to the flow of gradients across time. This is possible because of the discrete addressing used for read and write operations. 

However, the main challenge for the model is to learn proper read and write mechanisms so that it can write the hidden states of the previous time steps that will be useful for future predictions and read them at the right time step. We call this the reader/writer synchronization problem. Instead of designing complicated addressing mechanisms to mitigate the difficulty of learning how to properly address the external memory, TARDIS side-steps the reader/writer synchronization problem by using the following heuristics. For the first $k$ time steps, our model writes the micro-states into the $k$ cells of the memory in a sequential order. When the memory becomes full, the most effective strategy in terms of preserving the information stored in the memory would be to replace the memory cell that has been read with the micro-state generated from the hidden state of the controller after it is conditioned on the memory cell that has been read. If the model needs to perfectly retain the memory cell that it has just overwritten, the controller can in principle learn to do that by copying its read input to its write output (into the same memory cell). The pseudocode and the details of the memory update algorithm for TARDIS is presented in Algorithm \ref{algo:mem_update}.

\begin{algorithm}[ht]
\caption{Pseudocode for the controller and memory update mechanism of TARDIS.}
\begin{algorithmic}
    \STATE Initialize $\vh_0$
    \STATE Initialize $\mM_0$
    \FOR {$t \in \{1, \cdots T_x\}$}
        \STATE Compute the read weights $\overline{\vw}^r_t \leftarrow \text{read}(\vh_t, \mM_t, \vx_t)$
        \STATE Sample from/discretize $\overline{\vw}^r_t$ and obtain $\vw^r_t$
        \STATE Read from the memory, $\vr_t \leftarrow (\mM_t)^{\top}\vw^r_t$.
        \STATE Compute a new controller hidden state, $\vh_i \leftarrow \phi(\vx_t, \vh_{t-1}, \vr_t)$
        \IF {$t \le k$}            
            \STATE Write into the memory, $\mM_t[t] \leftarrow \mW_m \vh_t$
        \ELSE
            \STATE Select the memory location to write into $j \leftarrow \text{max}_j (\vw^r_t[j])$
            \STATE Write into the memory, $\mM_t[j] \leftarrow \mW_m \vh_t$
        \ENDIF
    \ENDFOR
\end{algorithmic}
\label{algo:mem_update}
\end{algorithm}

There are two missing pieces in Algorithm \ref{algo:mem_update}: How to generate the read weights? What is the structure of the controller function $\phi$? We will answer these two questions in detail in next two sub-sections.

\subsection{Addressing mechanism}

Similar to D-NTM, memory matrix $\mM_t$ of TARDIS has disjoint address section $\mA_t \in \R^{k \times a}$ and content section $\mC_t \in \R^{k \times c}$, $\mM_t = [\mA_t; \mC_t]$ and $\mM_t \in \R^{k \times q}$ for $q=c+a$. However, unlike D-NTM address vectors are fixed to random sparse vectors. The controller reads both the address and the content parts of the memory, but it will only write into the content section of the memory. 

The continuous read weights $\overline{\vw}^r_t$ are generated by an MLP which uses the information coming from $\vh_t$, $\vx_t$, $\mM_t$ and the usage vector $\vu_t$ (described below). The MLP is parametrized as follows:

\begin{align}
    \label{eqn:att_read}
    \pi_t[i] &= \va^{\top}\text{tanh}(\mW^{\gamma}_h \vh_t + \mW^{\gamma}_x \vx_t + \mW^{\gamma}_m \mM_t[i] + \mW_u^{\gamma} \vu_t) \\
    \overline{\vw}^r_t &= \text{softmax}(\pi_t),\label{gumblink}
\end{align}

where $\{\va, \mW^{\gamma}_h, \mW^{\gamma}_x, \mW^{\gamma}_m, \mW^{\gamma}_u\}$ are learnable parameters. $\vw^r_t$ is a one-hot vector obtained by either sampling from $\overline{\vw}^r_t$ or by using argmax over $\overline{\vw}^r_t$.

$\vu_t$ is the usage vector which denotes the frequency of accesses to each cell in the memory. $\vu_t$ is computed from the sum of discrete address vectors $\vw^r_t$ and normalizing them.

\begin{equation}
    \label{eqn:norm_usages}
    \vu_t = \text{norm}(\sum_{i=1}^{t-1} \vw^r_i).
\end{equation}

$\text{norm}(\cdot)$ applied in Equation \ref{eqn:norm_usages} is a simple feature-wise computation of centering and divisive variance normalization. This normalization step makes the training easier with the usage vectors. The introduction of the usage vector can help the attention mechanism to choose between the different memory cells based on their frequency of accesses to each cell of the memory. For example, if a memory cell is very rarely accessed by the controller, for the next time step, it can learn to assign more weights to those memory cells by looking into the usage vector. By this way, the controller can learn an LRU access mechanism \citep{santoro2016one,gulcehre2016dynamic}.

Further, in order to prevent the model to learn deficient addressing mechanisms, for e.g. reading the same memory cell which will not increase the memory capacity of the model, we decrease the probability of the last read memory location by subtracting $100$ from the logit of $\overline{\vw}^r_t$ for that particular memory location.

\subsection{TARDIS Controller}

We use an LSTM controller, and its gates are modified to take into account the content $\vr_t$ of the cell read from the memory:
\begin{equation}
\label{eqn:lstm_gates_tardis}
  \begin{pmatrix}
  {\ff}_t \\
  {\ffi}_t \\
  {\fo}_t 
  \end{pmatrix}
  =
  \begin{pmatrix}
  \mathtt{sigm}\\
  \mathtt{sigm}\\
  \mathtt{sigm}\\
  \end{pmatrix}  
    \left( \mW_h \vh_{t-1}~+~\mW_x \vx_{t}~+~\mW_r \vr_{t} \right), 
\end{equation}

where ${\ff}_t$, ${\ffi}_t$, and ${\fo}_t$ are forget gate, input gate, and output gate respectively. $\alpha_t, \beta_t$~ are the scalar RESET gates which control the magnitude of the information flowing from the memory and the previous hidden states to the cell of the LSTM $\vc_t$.  By controlling the flow of information into the LSTM cell, those gates will allow the model to store the sub-sequences or chunks of sequences into the memory instead of the entire context.

We use Gumbel sigmoid~\citep{maddison2016concrete,jang2016categorical} for $\alpha_t$ and $\beta_t$ due to its behavior close to binary. 

\begin{equation}
    \label{eqn:reset_alpha}
    \begin{pmatrix} \alpha_t \\
    \beta_t \\ 
    \end{pmatrix} = \begin{pmatrix} 
                \text{gumbel-sigmoid}\\
                \text{gumbel-sigmoid}\\
                \end{pmatrix} 
                \begin{pmatrix}
                \begin{pmatrix} \vw_h^{\alpha\top}\\ \vw_h^{\beta\top}\\ \end{pmatrix}\vh_{t-1} +  \begin{pmatrix} \vw_x^{\alpha\top} \\  \vw_x^{\beta \top} \\\end{pmatrix}\vx_t +        
                \begin{pmatrix} \vw_r^{\alpha\top}\\ \vw_r^{\beta\top}\\ \end{pmatrix} \vr_t
\end{pmatrix},
\end{equation}


As in Equation \ref{eqn:reset_alpha} empirically, we find gumbel-sigmoid to be  easier to train than the regular sigmoid. The temperature of the Gumbel-sigmoid is fixed to $0.3$ in all our experiments.

The cell of the LSTM controller, $\vc_t$ is computed according to the Equation \ref{eqn:lstm_cell_gate} with the $\alpha_t$ and $\beta_t$ RESET gates. 

\begin{align}
    \label{eqn:lstm_cell_gate}
        \tilde{\vc}_t &= \tanh(\beta_t \mW^{\fg}_h \vh_{t-1} + \mW^{\fg}_x \vx_{t} + \alpha_t \mW^{\fg}_r \vr_{t}), \nonumber\\
        \vc_t &= \ff_t \vc_{t-1} + \ffi_t \tilde{\vc}_t,
\end{align}

The hidden state of the LSTM controller is computed as follows:

\begin{equation}
\vh_t = \fo_t \tanh(\vc_t).
\end{equation}

\begin{figure}[h]
\begin{center}
\includegraphics[scale=0.4]{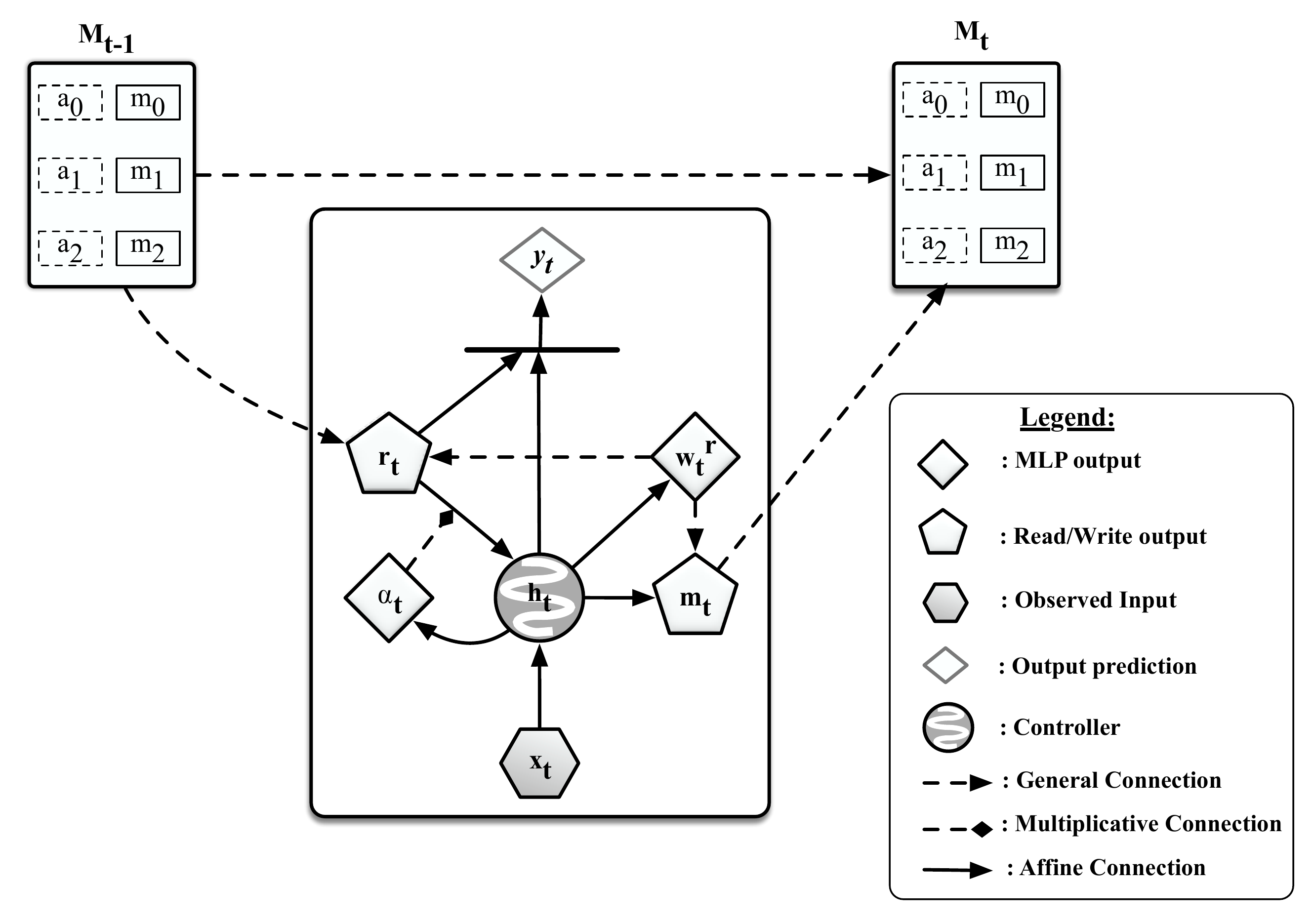}
\end{center}
\caption{At each time step controller takes $\vx_t$, the memory cell that has been read $\vr_t$ and the hidden state of the previous timestep $\vh_{t-1}$. Then, it generates $\alpha_t$ which controls the contribution of the $\vr_t$ into the internal dynamics of the new controller's state $\vh_t$ (We omit the $\beta_t$ in this visualization). Once the memory $\mM_t$ becomes full, discrete addressing weights $\vw^r_t$ is generated by the controller which will be used to both read from and write into the memory. To the predict the target $\vy_t$, the model will have to use both $\vh_t$ and $\vr_t$. }
\label{fig:tardis_overviewv2}
\end{figure}

In Figure \ref{fig:tardis_overviewv2}, we illustrate the interaction between the controller and the memory with various heads and components of the controller. 

\subsection{Micro-states and Long-term Dependencies}

A micro-state of the LSTM for a particular time step is the summary of the information that has been stored in the LSTM controller of the model. By attending over the cells of the memory which contains previous micro-states of the LSTM, the model can explicitly learn to restore information from its own past.

\begin{figure}[h]
\centering

\includegraphics[scale=0.3]{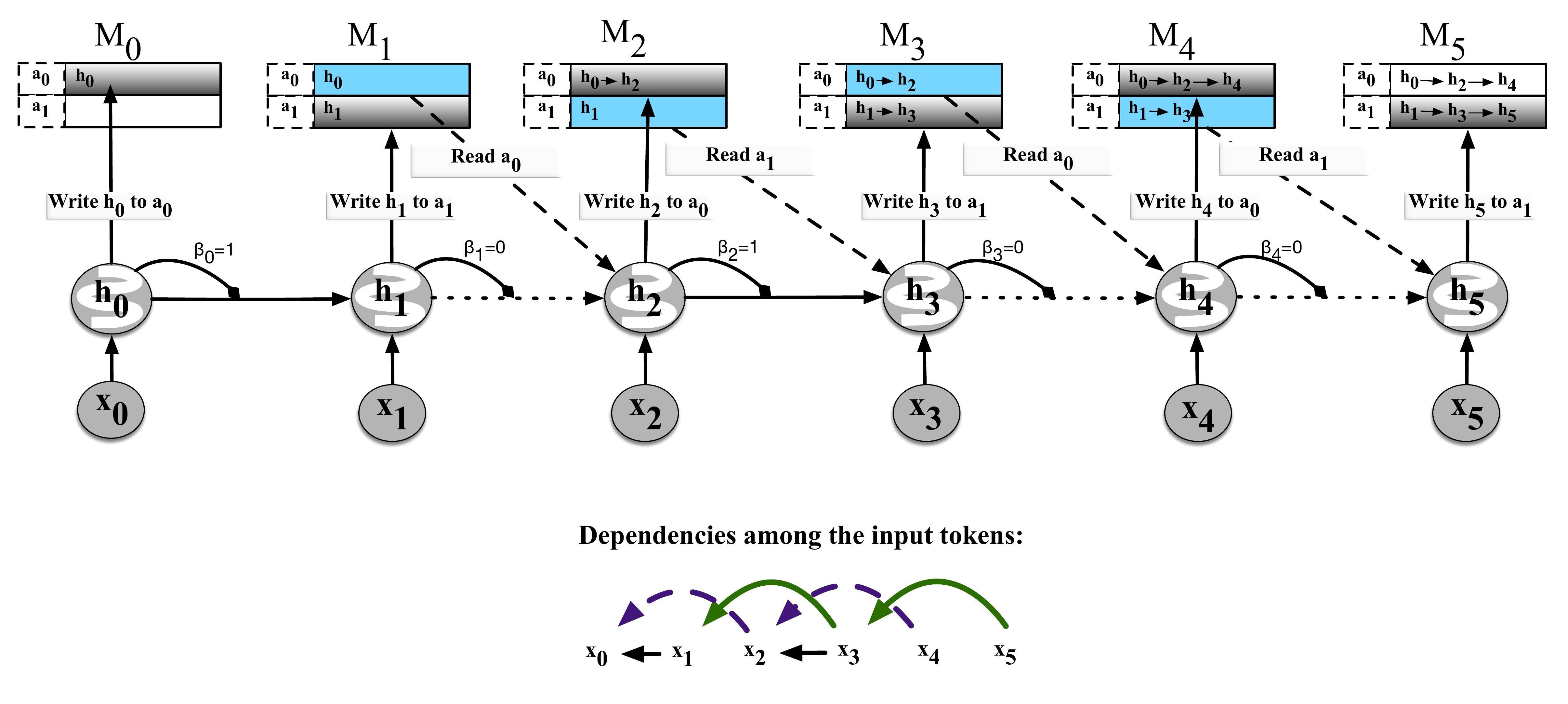}

\caption{TARDIS's controller can learn to represent the dependencies among the inputs tokens by choosing which cells to read and write and creating wormhole connections. $\vx_t$ represents the input to the controller at timestep $t$ and the $\vh_t$ is the hidden state of the controller RNN. }
\label{fig:tardis_overview_simple}
\end{figure}

The controller can learn to represent high-level temporal abstractions by creating wormhole connections through the memory as illustrated in Figure \ref{fig:tardis_overview_simple}. In this example, the model takes the token $x_0$ at the first timestep and stores its representation to the first memory cell with address $a_0$. In the second timestep, the controller takes $x_1$ as input and writes into the second memory cell with the address $a_1$. Furthermore, $\beta_1$ gater blocks the connection from $\vh_1$ to $\vh_2$.  At the third timestep, the controller starts reading. It receives $x_2$ as input and reads the first memory cell where micro-state of $\vh_0$ was stored. After reading, it computes the hidden-state $\vh_2$ and writes the micro-state of $\vh_2$ into the first memory cell. The length of the path passing through the microstates of $\vh_0$ and $\vh_2$  would be $1$. The wormhole connection from $\vh_2$ to $\vh_0$ would skip a timestep.

A regular single-layer RNN has a fixed graphical representation of a linear-chain when considering only the connections through its recurrent states or the temporal axis. However, TARDIS is more flexible in terms of that and it can learn directed graphs with more diverse structures using the wormhole connections and the RESET gates. The directed graph that TARDIS can learn through its recurrent states have at most the degree of 4 at each vertex (maximum 2 incoming and 2 outgoing edges) and it depends on the number of cells ($k$) that can be stored in the memory.

In this work, we focus on a variation of TARDIS, where the controller maintains a fixed-size external memory. However as in \citep{cheng2016long}, it is possible to use a memory that grows with respect to the length of its input sequences, but that would not scale and can be more difficult to train with discrete addressing.

\section{Training TARDIS}

In this section, we explain how to train TARDIS as a language model. We use language modeling as an example application. However, we would like to highlight that TARDIS can also be applied to any complex sequence to sequence learning tasks.

Consider $N$ training examples where each example is a sequence of length $T$. At every time-step $t$, the model receives the input $\vx_t \in \{0,1\}^{|V|}$ which is a one-hot vector of size equal to the size of the vocabulary $|V|$ and should produce the output $\vy_t \in \{0,1\}^{|V|}$ which is also a one-hot vector of size equal to the size of the vocabulary $|V|$.

The output of the model for $i$-th example and $t$-th time-step is computed as follows:

\begin{equation}
    \label{eqn:prediction}
    \vo_t^i =  \text{softmax}(\mW^o\fg(\vh^{(i)}_t,\vr^{(i)}_t)),
\end{equation}

where $\mW^o$ is the learnable parameters and $\fg(\vh_t, \vr_t)$ is a single layer MLP which combines both $\vh_t$ and $\vr_t$ as in deep fusion by \citep{pascanu2013construct}. The task loss would be the categorical cross-entropy between the targets and model-outputs. Super-script $i$ denotes that the variable is the output for the $i^{th}$ sample in the training set. 

\begin{equation}
    \label{eqn:prediction_cost_reinforce}
    \LL_{\text{model}}(\TT) =  -\frac{1}{N} \sum_{i=1}^N \sum_{t=1}^T \sum_{k=1}^{|V|} \vy_t^{(i)}[k] \log(o_t^{(i)}[k]),
\end{equation}

However, the discrete decisions taken for memory access during every time-step makes the model not differentiable and hence we need to rely on approximate methods of computing gradients with respect to the discrete address vectors. In this paper we explore two such approaches: REINFORCE~\citep{williams92} and straight-through estimator~\citep{bengio2013estimating}.

 \subsection{Using REINFORCE}
REINFORCE is a likelihood-ratio method, which provides a convenient and simple way of estimating the gradients of the stochastic actions. In this paper, we focus on application of REINFORCE on sequential prediction tasks, such as language modelling. For example $i$, let $R(\vw_j^{r(i)})$ be the reward for the action $\vw_j^{r(i)}$ at timestep $j$. We are interested in maximizing the expected return for the whole episode as defined below:

\begin{equation}
    \label{eqn:reinf_objective}
    \mathcal{J}(\TT) = \E[\sum_{j=0}^{T}R(\vw_j^{r(i)})]
\end{equation}

Ideally we would like to compute the gradients for Equation \ref{eqn:reinf_objective}, however computing the gradient of the expectation may not be feasible. We would have to use a Monte-Carlo approximation and compute the gradients by using the REINFORCE for the sequential prediction task which can be written as in Equation \ref{eqn:seq_pred_reinf}.

\begin{equation}
    \label{eqn:seq_pred_reinf}
    \nabla_{\TT}\mathcal{J}(\TT) = \frac{1}{N} \sum_{i=1}^N [\sum_{j=0}^{T}(R(\vw_j^{r(i)}) - b_j)\sum_{t=0}^{T}\nabla_{\TT}\log(\overline{\vw}^{r(i)}_t)],
\end{equation}

where $b_j$ is the reward baseline. However, we can further assume that the future actions do not depend on the past rewards in the episode/trajectory and further reduce the variance of REINFORCE as in Equation \ref{eqn:var_reduced_reinf}.
\begin{equation}
    \label{eqn:var_reduced_reinf}
    \nabla_{\TT}\mathcal{J}(\TT) =  \frac{1}{N} \sum_{i=1}^N [\sum_{t=0}^{T}\sum_{j=t}^T( R(\vw_j^{r(i)}) - b_j)\nabla_{\TT}\log(\overline{\vw}^{r(i)}_t)],
\end{equation}

In our preliminary experiments, we find out that the training of the model is easier with the discounted returns, instead of using the centered undiscounted return:
\begin{equation}
    \label{eqn:var_reduced_reinf_disc}
    \nabla_{\TT}\mathcal{J}(\TT) =  \frac{1}{N} \sum_{i=1}^N [\sum_{t=0}^{T}\sum_{j=t}^T[\gamma^{j-t}( R(\vw_j^{r(i)}) - b_j)]\nabla_{\TT}\log(\overline{\vw}^{r(i)}_t)].
\end{equation}
\paragraph{Training REINFORCE with an Auxiliary Cost} Training models with REINFORCE can be difficult, due to the variance imposed into the gradients. In the recent years, researchers have developed several tricks in order to mitigate the effect of high-variance in the gradients. As proposed by \citep{mnih2014neural}, we also use variance normalization on the REINFORCE gradients. 

For TARDIS, reward at timestep $j$ ($R(\vw_j^{r(i)})$) is the log-likelihood of the prediction at that timestep. Our initial experiments showed that REINFORCE with this reward structue often tends to under-utilize the memory and mainly rely on the internal memory of the LSTM controller. Especially, in the beginning of the training model, it can just decrease the loss by relying on the memory of the controller and this can cause the REINFORCE to increase the log-likelihood of the random actions.

In order to deal with this issue, instead of using the log-likelihood of the model as reward, we introduce an auxiliary cost to use as the reward $R'$ which is computed based on predictions which are only based on the memory cell $\vr_t$ which is read by the controller and not the hidden state of the controller:

\begin{equation}
    \label{eqn:auxilary_cost_reinforce}
    R'(\vw_j^{r(i)}) =  \sum_{k=1}^{|V|} \vy_j^{(i)}[k] \log(\text{softmax}(\mW_{r}^o\bar{\vr}^{(i)}_j + \mW_{x}^o\vx^{(i)}_j))[k],
\end{equation}

In Equation \ref{eqn:auxilary_cost_reinforce}, we only train the parameters $\{\mW_{r}^o \in \R^{d_o \times d_m}, \mW_{x}^o \in \R^{d_o \times d_x}\}$ where $d_o$ is the dimensionality of the output size and $d_x$ (for language modelling both $d_o$ and $d_x$ would be $d_o=|V|$) is the dimensionality of the input of the model. We do not backpropagate through $\vr_i^{(j)}$ and thus we denote it as $\bar{\vr}^{(j)}_i$ in our equations.


\subsection{Using Gumbel Softmax}
Training with REINFORCE can be challenging due to the high variance of the gradients, gumbel-softmax provides a good alternative with straight-through estimator for REINFORCE to tackle the variance issue. Unlike \citep{maddison2016concrete,jang2016categorical} instead of annealing the temperature or fixing it, our model learns the inverse-temperature with an MLP $\tau(\vh_t)$ which has a single scalar output conditioned on the hidden state of the controller. 
\begin{align}
    \tau(\vh_t) &= \text{softplus}(\vw^{\tau\top}\vh_t + \vb^{\tau}) + 1.\\
    \text{gumbel-softmax}(\pi_t[i]) &= \text{softmax}((\pi_t[i] + \xi) \tau(\vh_t)),
\end{align}

We replace the softmax in Equation \ref{gumblink} with gumbel-softmax defined above. During forward computation, we sample from $\overline{\vw}^r_t$ and use the generated one-hot vector ${\vw}^r_t$ for memory access. However, during backprop, we use $\overline{\vw}^r_t$ for gradient computation and hence the entire model becomes differentiable.

Learning the temperature of the Gumbel-Softmax reduces the burden of performing extensive hyper-parameter search for the temperature.

\section{Related Work}

Neural Turing Machine (NTM) \citep{graves2014neural} is the most related class of architecture to our model. NTMs have proven to be successful in terms of generalizing over longer sequences than the sequences that it has been trained on. Also NTM has been shown to be more effective in terms of solving algorithmic tasks than the gated models such as LSTMs. However NTM can have limitations due to some of its design choices. Due to the controller's lack of precise knowledge on the contents of the information, the contents of the memory can overlap. These memory augmented models are also known to be complicated, which yields to the difficulties in 
terms of implementing the model and training it. The controller has no information about the sequence of operations and the information such as frequency of the read and write access to the memory. TARDIS tries to address these issues.

\cite{gulcehre2016dynamic} proposed a variant of NTM called dynamic NTM (D-NTM) which had learnable location based addressing. D-NTM can be used with both continuous addressing and discrete addressing. Discrete D-NTM is related to TARDIS in the sense that both models use discrete addressing for all the memory operations. However, discrete D-NTM expects the controller to learn to read/write and also learn reader/writer synchronization. TARDIS do not have this synchronization problem since both reader and writer are tied. \cite{rae2016scaling} proposed sparse access memory (SAM) mechanism for NTMs which can be seen as a hybrid of continuous and discrete addressing. SAM uses continuous addressing over a selected set of top-$K$ relevant memory cells. Recently, \cite{Graves_Nature2016} proposed a differentiable neural computer (DNC) which is a successor of NTM.

\cite{rocktaschel2015reasoning} and \citep{cheng2016long} proposed models that generate weights to attend over the previous hidden states of the RNN. However, since those models attend over the whole context, the computation of the attention can be inefficient. 

\cite{grefenstette2015learning} has proposed a model that can store the information in a data structure, such as in a stack, dequeue or queue in a differentiable manner.

\cite{grave2016improving} has proposed to use a cache based memory representation which stores the last $k$ states of the RNN in the memory and similar to the traditional cache-based models the model learns to choose a state of the memory for the prediction in the language modeling tasks \citep{kuhn1990cache}.

\section{Gradient Flow through the External Memory}
\label{sec:gradient_flow}
In this section, we analyze the flow of the gradients through the external memory and will also investigate its efficiency in terms of dealing with the vanishing gradients problem \citep{hochreiter1991untersuchungen,bengio1994learning}. First, we describe the vanishing gradient problem in an RNN and then describe how an external memory model can deal with it. For the sake of simplicity, we will focus on vanilla RNNs during the entire analysis, but the same analysis can be extended to LSTMs. In our analysis, we also assume that the weights for the read/write heads are discrete. 

We will show that the rate of the gradients vanishing through time for a memory-augmented recurrent neural network is much smaller than of a regular vanilla recurrent neural network.

Consider an RNN which at each timestep $t$ takes an input $\vx_t \in \R^d$ and produces an output $\vy_t \in \R^o$. The hidden state of the RNN can be written as,

\begin{align}
    \vz_t &= \mW \vh_{t-1} + \mU \vx_t,  \\
    \vh_t &= \ff(\vz_t).
\end{align}

where $\mW$ and $\mU$ are the recurrent and the input weights of the RNN respectively and $\ff(\cdot)$ is a non-linear activation function. Let $\LL=\sum_{t=1}^T \LL_t$ be the loss function that the RNN is trying to minimize. Given an input sequence of length $T$, we can write the derivative of the loss $\LL$ with respect to parameters $\TT$ as,

\begin{equation}
    \frac{\partial\LL}{\partial\TT}~=~\sum_{1 \le t_1 \le T} \frac{\partial\LL_{t_1}}{\partial\TT}~=~\sum_{1 \le t_1 \le T}
    \sum_{1 \le t_0 \le t_1}\frac{\partial\LL_{t_1}}{\partial\vh_{t_1}}\frac{\partial\vh_{t_1}}{\partial\vh_{t_0}}\frac{\partial\vh_{t_0}}{\partial\TT}.
\end{equation}

The multiplication of many Jacobians in the form of $\frac{\partial\vh_t}{\partial\vh_{t-1}}$ to obtain $\frac{\partial \vh_{t_1}}{\partial \vh_{t_0}}$ is the main reason of the vanishing and the exploding gradients \citep{pascanu2013difficulty}:

\begin{equation}
    \frac{\partial \vh_{t_1}}{\partial \vh_{t_0}} = \prod_{t_0 < t \le t_1} \frac{\partial \vh_t}{\partial \vh_{t-1}} = \prod_{t_0 < t
        \le t_1} \text{diag}[\ff^{\prime}(\vz_t)] \mW,
\end{equation}

Let us assume that the singular values of a matrix $\mM$ are ordered as, $\sigma_1({\mM}) \ge \sigma_2({\mM}) \ge \dots \ge \sigma_n({\mM})$. Let $\alpha$ be an upper bound on the singular values of $\mW$, s.t. $\alpha \ge  \sigma_{1}(\mW)$, then the norm of the Jacobian will satisfy \citep{zilly2016recurrent},

\begin{equation}
    ||\frac{\partial \vh_t}{\partial \vh_{t-1}}|| \le ||\mW|| ~~ ||\text{diag}[\ff^{\prime}(\vz_t)|| \le \alpha~ \sigma_{1}(\text{diag}[\ff^{\prime}(\vz_t)]),
\end{equation}

\cite{pascanu2013difficulty} showed that for $||\frac{\partial\vh_t}{\partial\vh_{t-1}}|| \le \sigma_1(\frac{\partial\vh_t}{\partial\vh_{t-1}})\le \eta < 1$, the following inequality holds:

\begin{equation}
    \label{eqn:bound_sing_val_ineq}
    ||\prod_{t_0 \le t \le t_1} \frac{\partial \vh_t}{\partial \vh_{t-1}}|| \le \sigma_1\left(\prod_{t_0 \le t \le t_1} \frac{\partial \vh_t}{\partial \vh_{t-1}}\right) \le \eta^{t_1 - t_0} .    
\end{equation}

Since $\eta < 1$ and the norm of the product of Jacobians grows exponentially on $t_1 - t_0$, the norm of the gradients will vanish exponentially fast.

Now consider the MANN where the contents of the memory are linear projections of the previous hidden states as described in Equation~\ref{eqn:linproj}. Let us assume that both reading and writing operation use discrete addressing. Let the content read from the memory at time step $t$ correspond to some memory location $i$:

\begin{equation}
\vr_t = \mM_t[i] = \mA \vh_{i_t},
\end{equation}

 where $\vh_{i_t}$ corresponds to the hidden state of the controller at some previous timestep $i_t$.

Now the hidden state of the controller in the external memory model can be written as,

\begin{align}
    \label{eqn:mem_aug_hids}
    \vz_t &= \mW\vh_{t-1} + \mV\vr_t + \mU\vx_t \nonumber,\\
    \vh_t &= \ff(\vz_t).
\end{align}

If the controller reads $\mM_t[i]$ at time step $t$ and its memory content is $\mA\vh_{i_t}$ as described above, then the Jacobians associated with Equation \ref{eqn:mem_aug_hids} can be computed as follows:

\begin{align}
\frac{\partial \vh_{t_1}}{\partial \vh_{t_0}} &= \prod_{t_0 < t \le t_1} \frac{\partial \vh_t}{\partial \vh_{t-1}} \nonumber\\
& = \left(\prod_{t_0 < t
        \le t_1} \text{diag}[\ff^{\prime}(\vz_t)] \mW \right)  + \sum_{k=t_0}^{t_1-1}(\prod_{k < t^{\ast} < t_1} \text{diag}[\ff^{\prime}(\vz_{t^{\ast}})]\mW)~\text{diag}[\ff^{\prime}(\vz_{k})] \mV \mA \frac{\partial \vh_{i_k}}{\partial \vh_{t_0}}\nonumber\\
        &~~~~~~~~~~~~~~~~~~~~~~~~~~~~~~~~~~~~~~~~~~~~~~~~~~~~~~~~~+~\text{diag}[\ff^{\prime}(\vz_{t_1})] \mV \mA \frac{\partial \vh_{i_{t_1}}}{\partial \vh_{t_0}}\\
& = \mQ_{t_1 t_0} + \mR_{t_1 t_0}\label{eqn:mem_augmented_jb}.\\\nonumber
\end{align}

where $\mQ_{t_1 t_0}$ and $\mR_{t_1 t_0}$ are defined as below,
\begin{align}
\label{eqn:q_r_defs}
\mQ_{t_1 t_0} &= \prod_{t_0 < t \le t_1} \text{diag}[\ff^{\prime}(\vz_t)] \mW, \\
\mR_{t_1 t_0} &= \sum_{k=t_0}^{t-1}(\prod_{k < t^{\ast} < t} \text{diag}[\ff^{\prime}(\vz_{t^{\ast}})]\mW)~\text{diag}[\ff^{\prime}(\vz_{k})] \mV \mA \frac{\partial \vh_{i_k}}{\partial \vh_{t_0}} + \text{diag}[\ff^{\prime}(\vz_{t_1})] \mV \mA \frac{\partial \vh_{i_{t_1}}}{\partial \vh_{t_0}} .
\end{align}

As shown in Equation \ref{eqn:mem_augmented_jb}, Jacobians of the MANN can be rewritten as a summation of two matrices, $\mQ_{t_1 t_0}$ and $\mR_{t_1 t_0}$. The gradients flowing through $\mR_{t_1 t_0}$ do not necessarily vanish through time, because it is the sum of jacobians computed over the shorter paths. 

The norm of the Jacobian can be lower bounded as follows by using Minkowski inequality:
\begin{align}
|| \frac{\partial \vh_{t_1}}{\partial \vh_{t_0}}|| &= ||\prod_{t_0 < t \le t_1} \frac{\partial \vh_t}{\partial \vh_{t-1}}||\\
& = ||\mQ_{t_1 t_0} + \mR_{t_1 t_0} || \ge  ||\mR_{t_1 t_0}|| - ||\mQ_{t_1 t_0}||
\end{align}

Assuming that the length of the dependency is very long $||\mQ_{t_1 t_0}||$ would vanish to 0. Then we will have,
\begin{equation}
    \label{eqn:simple_vanishing_lower_qr}
     ||\mQ_{t_1 t_0} + \mR_{t_1 t_0} || \ge ||\mR_{t_1 t_0}|| 
\end{equation}

As one can see that the rate of the gradients vanishing through time depends on the length of the sequence passes through $\mR_{t_1 t_0}$. This is typically lesser than the length of the sequence passing through $\mQ_{t_1 t_0}$. Thus the gradients vanish at lesser rate than in an RNN. In particular the rate would strictly depend on the length of the shortest paths from $t_1$ to $t_0$, because for the long enough dependencies, gradients through the longer paths would still vanish.

We can also derive an upper bound for norm of the Jacobian as follows:
\begin{align}
|| \frac{\partial \vh_{t_1}}{\partial \vh_{t_0}}|| &= ||\prod_{t_0 < t \le t_1} \frac{\partial \vh_t}{\partial \vh_{t-1}}||\\
& = ||\mQ_{t_1 t_0} + \mR_{t_1 t_0} || \le  \sigma_1(\mQ_{t_1 t_0} + \mR_{t_1 t_0})
\end{align}

Using the result from \citep{loyka2015singular}, we can lower bound $\sigma_1(\mQ_{t_1 t_0} + \mR_{t_1 t_0})$ as follows:
\begin{equation}
\label{eqn:singular_value_prop}
\sigma_1(\mQ_{t_1 t_0} + \mR_{t_1 t_0}) \ge |\sigma_1(\mQ_{t_1 t_0}) - \sigma_1(\mR_{t_1 t_0})|  
\end{equation}

For long sequences we know that $\sigma_1(\mQ_{t_1 t_0})$ will go to $0$ (see equation \ref{eqn:bound_sing_val_ineq}). Hence,
\begin{equation}
 \label{eqn:vanishing_mem_aug2}
 \sigma_1(\mQ_{t_1 t_0} + \mR_{t_1 t_0}) \ge \sigma_1(\mR_{t_1 t_0}) 
\end{equation}

The rate at which $\sigma_1(\mR_{t_1 t_0})$ reaches zero is strictly smaller than the rate at which $\sigma_1(\mQ_{t_1 t_0})$ reaches  zero and with ideal memory access, it will not reach zero. Hence unlike vanilla RNNs, Equation \ref{eqn:vanishing_mem_aug2} states that the upper bound of the norm of the Jacobian will not reach to zero for a MANN with ideal memory access.


\begin{theorem} 
    \label{theo:vanish_grad_perfect}
    Consider a memory augmented neural network with $T$ memory cells for a sequence of length $T$, and each hidden state of the controller is stored in different cells of the memory. If the prediction at time step $t_1$ has only a long-term dependency to $t_0$ and the prediction at $t_1$ is independent from the tokens appear before $t_0$, and the memory reading mechanism is perfect, the model will not suffer from vanishing gradients when we back-propagate from $t_1$ to $t_0$.\footnote{Let us note that, unlike an Markovian $n$-gram assumption, here we assume that at each time step the $n$ can be different.} 
\end{theorem}

\paragraph{Proof:}

If the input sequence has a longest-dependency to $t_0$ from $t_1$, we would only be interested in gradients propagating from $t_1$ to $t_0$ and the Jacobians from $t_1$ to $t_0$, i.e. $\frac{\partial\vh_{t_1}}{\partial\vh_{t_0}}$. If the controller learns a perfect reading mechanism at time step $t_1$ it would read memory cell where the hidden state of the RNN at time step $t_0$ is stored at. Thus following the jacobians defined in the Equation \ref{eqn:mem_augmented_jb}, we can rewrite the jacobians as,

\begin{align}
\frac{\partial \vh_{t_1}}{\partial \vh_{t_0}} &= \prod_{t_0 < t \le t_1} \frac{\partial \vh_t}{\partial \vh_{t-1}} \nonumber\\
& = \left(\prod_{t_0 < t
        \le t_1} \text{diag}[\ff^{\prime}(\vz_t)] \mW \right)  + \sum_{k=t_0}^{t_1-1}(\prod_{k < t^{\ast} < t_1} \text{diag}[\ff^{\prime}(\vz_{t^{\ast}})]\mW)~\text{diag}[\ff^{\prime}(\vz_{k})] \mV \mA \frac{\partial \vh_{i_k}}{\partial \vh_{t_0}}\nonumber\\
        &~~~~~~~~~~~~~~~~~~~~~~~~~~~~~~~~~~~~~~~~~~~~~~~~~~~~~~~~~+~\text{diag}[\ff^{\prime}(\vz_{t_1})] \mV \mA \frac{\partial \vh_{t_0}}{\partial \vh_{t_0}} \label{eqn:jacobian_products_opt}
\end{align}

In Equation \ref{eqn:jacobian_products_opt}, the first two terms might vanish as $t_1 - t_0$ grows. However, the singular values of the third term do not change as $t_1-t_0$ grows. As a result, the gradients propagated from $t_1$ to $t_0$ will not necessarily vanish through time. However, in order to obtain stable dynamics for the network, the initialization of the matrices, $\mV$ and $\mA$ is important. $\square$

This analysis highlights the fact that an external memory model with optimal read/write mechanism can handle long-range dependencies much better than an RNN. However, this is applicable only when we use discrete addressing for read/write operations. Both NTM and D-NTM still have to learn how to read and write from scratch which is a challenging optimization problem. For TARDIS tying the read/write operations make the learning to become much simpler for the model. In particular, the results of the Theorem \ref{theo:vanish_grad_perfect} points  the importance of coming up with better ways of designing attention mechanisms over the memory.

The controller of a MANN may not be able learn to use the memory efficiently. For example, some cells of the memory may remain empty or may never be read. The controller can overwrite the memory cells which have not been read. As a result the information stored in those overwritten memory cells can be lost completely. However TARDIS avoids most of these issues by the construction of the algorithm. 

\section{On the Length of the Paths Through the Wormhole Connections}
As we have discussed in Section \ref{sec:gradient_flow}, the rate at which the gradients vanish for a MANN depends on the length of the paths passing along the wormhole connections. In this section we will analyse those lengths in depth for untrained models such that the model will assign uniform probability to read or write all memory cells. This will give us a better idea on how each untrained model uses the memory at the beginning of the training.

A wormhole connection can be created by reading a memory cell and writing into the same cell in TARDIS. For example, in Figure \ref{fig:tardis_overview_simple}, while the actual path from $\vh_4$ to $\vh_0$ is of length 4, memory cell $\va_0$ creates a shorter path of length 2 $(\vh_0 \rightarrow \vh_2 \rightarrow \vh_4)$. We call the length of the actual path as $T$ and length of the shorter path created by wormhole connection as $T_{mem}$.

\begin{figure}[h]
\centering
\begin{minipage}{.5\textwidth}
\centering
\includegraphics[scale=0.35]{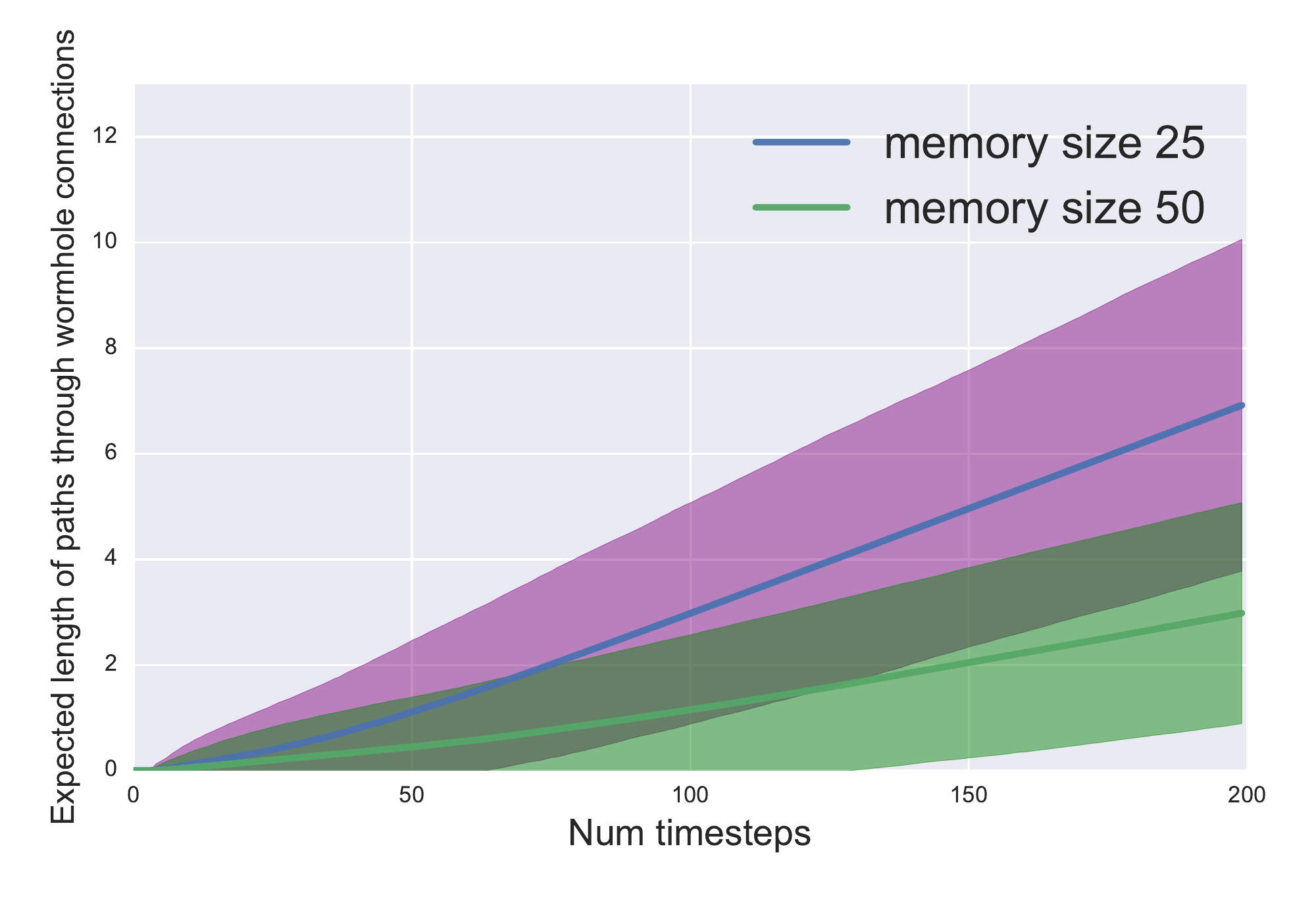} 
a)
\end{minipage}
\begin{minipage}{.45\textwidth}
\centering
\includegraphics[scale=0.35]{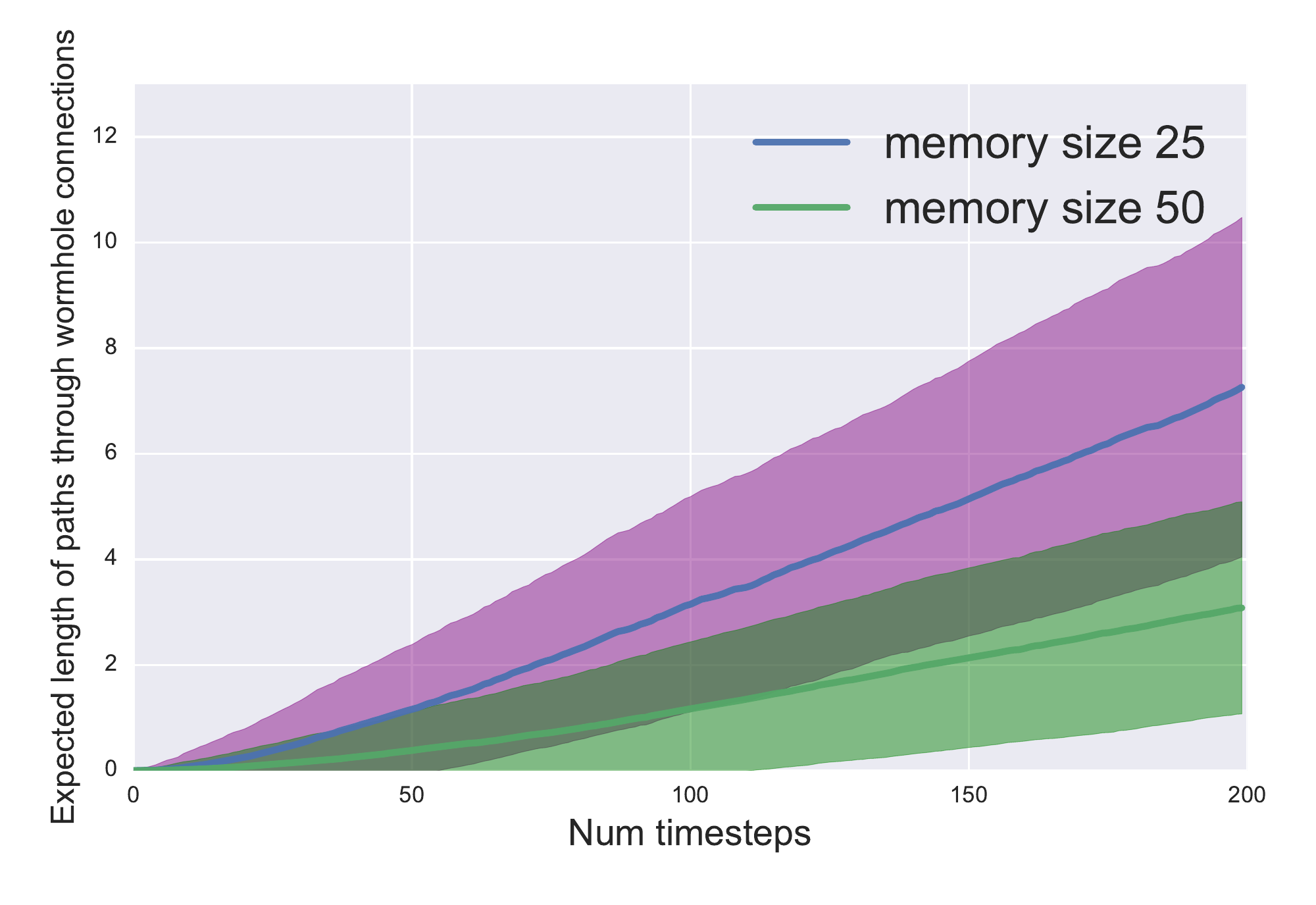}
b)
\end{minipage}

\caption{In these figures we visualized the expected path length in the memory cells for a sequence of length $200$, memory size $50$ with $100$ simulations. a) shows the results for the TARDIS and b) shows the simulation for a MANN with uniformly random read and write heads.}
\label{fig:mem_path_len}
\end{figure}

Consider a TARDIS model which has $k$ cells in its memory. If TARDIS access each memory cell uniformly random, then the probability of accessing a random cell $i$, $p[i]~=~\frac{1}{k}$.
The expected length of the shorter path created by wormhole connections ($T_{mem}$) would be proportional to the number of reads and writes into a memory cell. For TARDIS with reader choosing a memory cell uniformly random this would be $T_{mem} = \sum_{i=k}^T p[i]~=~\frac{T}{k} - 1$ at the end of the sequence. We verify this result by simulating the read and write heads of TARDIS as in Figure \ref{fig:mem_path_len} a).

Now consider a MANN with separate read and write heads each accessing the memory in discrete and uniformly random fashion. Let us call it as uMANN. We will compute the expected length  of the shorter path created by wormhole connections ($T_{mem}$) for uMANN. $\vw^r_t$ and $\vw^w_t$ are the read and write head weights, each sampled from a multinomial distribution with uniform probability for each memory cells respectively. Let $j_t$ be the index of the memory cell read at timestep $t$. For any memory cell $i$, $\text{len}(\cdot)$, defined below, is a recursive function that computes the length of the path created by wormhole connections in that cell.

\begin{equation}
 \label{eqn:comp_len_mem}
     \text{len}(\mM_t[i], i, j_t) = \left\{\begin{matrix}
\text{len}(\mM_{t-1}[j_t], i, j_t) + 1 & if~\vw^w_t[i] = 1\\ 
\text{len}(\mM_{t-1}[i], i, j_t) & if~\vw^w_t[i] = 0
\end{matrix}\right.
\end{equation}
 

It is possible to prove that $T_{mem}=\sum_t \E_{i,j_t}[\text{len}(\mM_t[i], i, j_t)]$ will be $T/k - 1$ by induction for every memory cell. However, for proof assumes that when $t$ is less than or equal to $k$, the length of all paths stored in the memory $\text{len}(\mM_t[i],~i,~j_t)$ should be $0$. We have run simulations to compute the expected path length in a memory cell of uMANN as in Figure \ref{fig:mem_path_len} (b).

This analysis shows that while TARDIS with uniform read head maintains the same expected length of the shorter path created by wormhole connections as uMANN, it completely avoids the reader/writer synchronization problem.

If $k$ is large enough, $T_{mem}<<T$ should hold. In expectation, $\sigma_1(\mR_{t_1 t_0})$ will decay proportionally to $T_{mem}$ whereas $\sigma_1(\mQ_{t_1 t_0})$ will decay proportional \footnote{Exponentially when the Equation \ref{eqn:bound_sing_val_ineq} holds.} to $T$.  With ideal memory access, the rate at which $\sigma_1(\mR_{t_1 t_0})$ reaches zero would be strictly smaller than the rate at which $\sigma_1(\mQ_{t_1 t_0})$ reaches zero. Hence, as per Equation \ref{eqn:vanishing_mem_aug2}, the upper bound of the norm of the Jacobian will vanish at a much smaller rate. However, this result assumes that the dependencies which the prediction relies are accessible through the memory cell which has been read by the controller.

\begin{figure}[h]
\begin{center}
\includegraphics[scale=0.35]{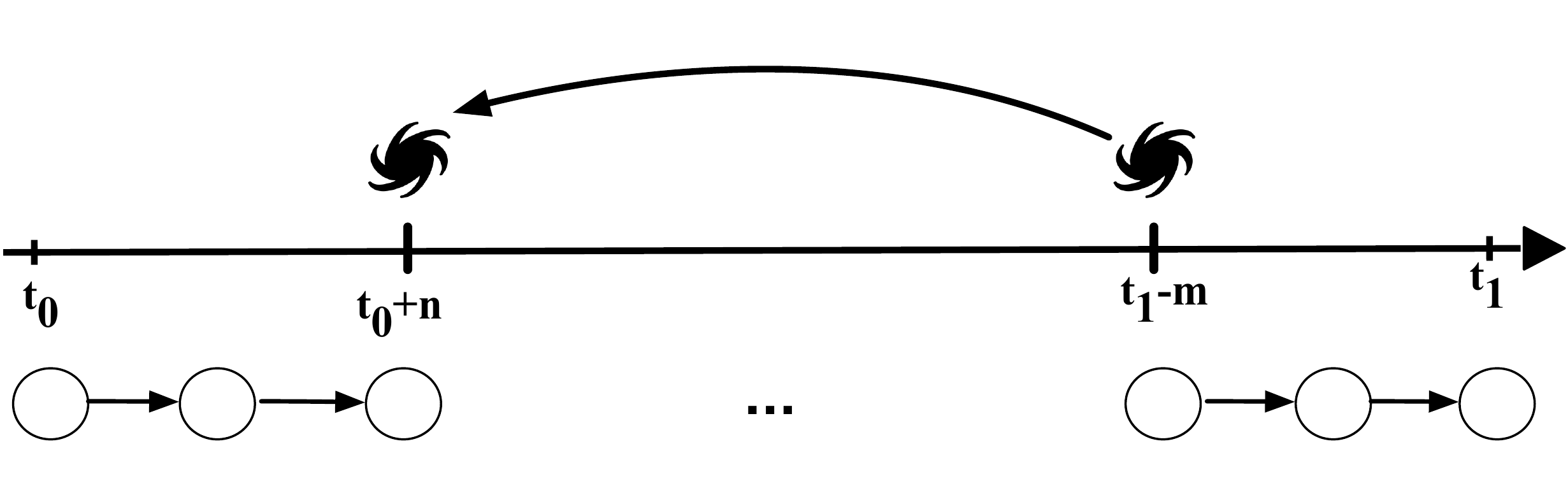}
\end{center}
\caption{Assuming that the prediction at $t_1$ depends on the $t_0$, a wormhole connection can shorten the path by creating a connection from $t_1-m$ to $t_0+n$. A wormhole connection may not directly create a connection from $t_1$ to $t_0$, but it can create shorter paths which the gradients can flow without vanishing. In this figure, we consider the case where a wormhole connection is created from $t_1-m$ to $t_0+n$. This connections skips all the tokens in between $t_1 -m$ and $t_0+n$.}
\label{fig:wormhole_conns}
\end{figure}


In the more general case, consider a MANN with $k \ge T$. The writer just fills in the memory cells in a sequential manner and the reader chooses a memory cell uniformly at random. Let us call this model as urMANN. Let us assume that there is a dependency between two timesteps $t_0$ and $t_1$ as shown in Figure \ref{fig:wormhole_conns}.  If $t_0$ was taken uniformly between $0$ and $t_1-1$, then there is a probability $0.5$ that the read address invoked at time $t_1$ will be greater than or equal to $t_0$ (proof by symmetry). In that case, the expected shortest path length through that wormhole connection would be $(t_1-t_0)/2$, but this still would not scale well. If the reader is very well trained, it could pick exactly $t_0$ and the path length will be 1.

Let us consider all the paths of length less than or equal to $k+1$ of the form in Figure \ref{fig:wormhole_conns}. Also, let $n \le k/2$ and $m \le k/2$. Then, the shortest path from $t_0$ to $t_1$ now has length $n+m+1 \le k+1$, using a wormhole connection that connects the state at $t_0+n$ with the state at $t_1-m$. There are $O(k^2)$ such paths that are realized, but we leave the distribution of the length of that shortest path as an open question. However, the probability of hitting a very short path (of length less than or equal to $k+1$) increases exponentially with $k$.  Let the probability of the read at $t_1-m$ to hit the interval $(t_0,~t_0~+~k/2)$ be $p$. Then the probability that the shorter paths over the last $k$ reads hits that interval is $1-(1-p)^{k/2}$, where $p$ is on the order of $k/t_1$. On the other hand, the probability of not hitting that interval approaches to 0 exponentially with $k$.

\begin{figure}[h]
\centering
\begin{minipage}{.44\textwidth}
\centering
\includegraphics[scale=0.34]{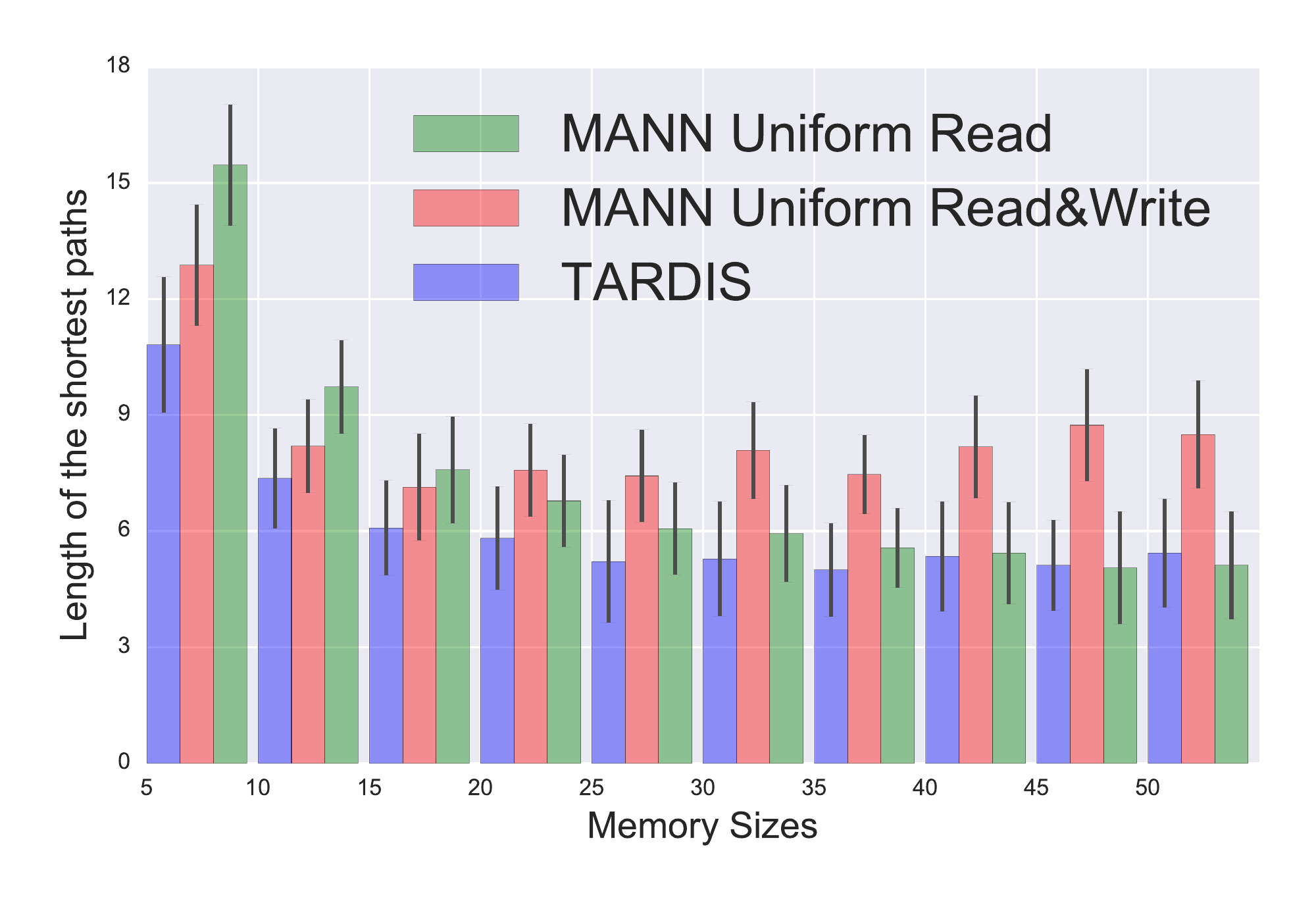} 
a)
\end{minipage}
\begin{minipage}{.44\textwidth}
\centering
\includegraphics[scale=0.34]{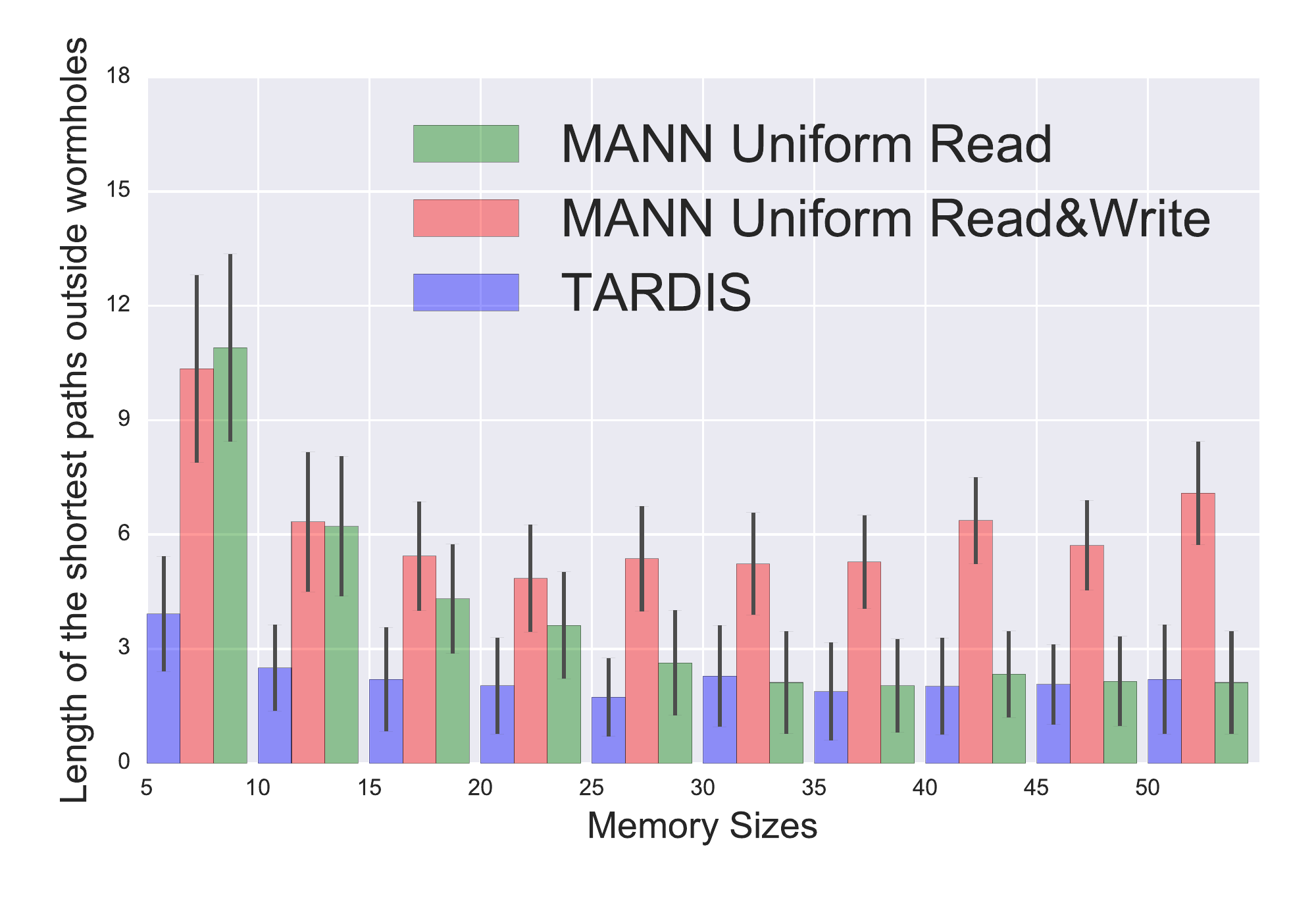}
b)
\end{minipage}
\caption{We have run simulations for TARDIS, MANN with uniform read and write mechanisms~(uMANN) and MANN with uniform read and write head is fixed with a heuristic~(urMANN). In our simulations, we assume that there is a dependency from timestep 50 to 5. We run 200 simulations for each one of them with different memory sizes for each model. In plot a) we show the results for the expected length of the shortest path from timestep 50 to 5. In the plots, as the size of the memory gets larger for both models, the length of the shortest path decreases dramatically. In plot b), we show the expected length of the shortest path travelled outside the wormhole connections with respect to different memory sizes. TARDIS seems to use the memory more efficiently compared to other models in particular when the size of the memory is small by creating shorter paths.}
\label{fig:bar_chart_exp_len}

\end{figure}

Figure \ref{fig:wormhole_conns} illustrates how wormhole connections can creater shorter paths. In Figure \ref{fig:bar_chart_exp_len} (b), we show that the expected length of the path travelled outside the wormhole connections obtained from the simulations decreases as the size of the memory decreases. In particular, for urMANN and TARDIS the trend is very close to exponential. As shown in Figure \ref{fig:bar_chart_exp_len} (a), this also influences the total length of the paths travelled from timestep 50 to 5 as well. Writing into the memory by using weights sampled with uniform probability for all memory cells can not use the memory as efficiently as other approaches that we compare to. In particular fixing the writing mechanism seems to be useful.

Even if the reader does not manage to learn where to read, there are many "short paths" which can considerably reduce the effect of vanishing gradients.

\section{On Generalization over the Longer Sequences}
\cite{graves2014neural} have shown that the LSTMs can not generalize well on the sequences longer than the ones seen during the training. Whereas a MANN such as an NTM or a D-NTM has been shown to generalize to sequences longer than the ones seen during the training set on a set of toy tasks. 

We believe that the main reason of why LSTMs typically do not generalize to the sequences longer than the ones that are seen during the training is mainly because the hidden state of an LSTM network utilizes an unbounded history of the input sequence and as a result, its parameters are optimized using the maximum likelihood criterion to fit on the sequences with lengths of the training examples. However, an n-gram language model or an HMM does not suffer from this issue. In comparison, an n-gram LM would use an input context with a fixed window size and an HMM has the Markov property in its latent space. As argued below, we claim that while being trained a MANN can also learn the ability to generalize for sequences with a longer length than the ones that appear in the training set by modifying the contents of the memory and reading from it.

A regular RNN will minimize the negative log-likelihood objective function for the targets $\vy_t$ by using the unbounded history represented with the hidden state of the RNN, and it will model the parametrized conditional distribution $p(\vy_t|\vh_t;\TT)$ for the prediction at timestep $t$ and a MANN would learn  $p(\vy_t|\vh_t, \vr_t; \TT)$. If we assume that $\vr_t$ represents all the dependencies that $\vy_t$ depends on in the input sequence, we will have  $p(\vy_t|\vh_t, \vr_t; \TT) \approx p(\vy_t|\vr_t, \vx_t;\TT)$ where $\vr_t$ represents the dependencies in a limited context window that only contains paths shorter than the sequences seen during the training set. Due to this property, we claim that MANNs such as NTM, D-NTM or TARDIS can generalize to the longer sequences more easily. In our experiments on PennTreebank, we show that a TARDIS language model trained to minimize the log-likelihood for $p(\vy_t|\vh_t, \vr_t; \TT)$ and on the test set  both $p(\vy_t|\vh_t, \vr_t; \TT)$ and $p(\vy_t|\vr_t, \vx_t; \TT)$ for the same model yields to very close results. On the other hand, the fact that the best results on bAbI dataset obtained in \citep{gulcehre2016dynamic} is with feedforward controller and similarly in \citep{graves2014neural} feedforward controller was used to solve some of the toy tasks also confirms our hypothesis. As a result, what has been written into the memory and what has been read becomes very important to be able to generalize to the longer sequences.
 
\section{Experiments}
\label{sec:exps}


\subsection{Character-level Language Modeling on PTB}
As a preliminary study on the performance of our model we consider character-level language modelling. We have evaluated our models on Penn TreeBank~(PTB) corpus \citep{marcus1993building} based on the train, valid and test used in \citep{mikolov2012subword}. On this task, we are using layer-normalization \citep{ba2016layer} and recurrent dropout \citep{semeniuta2016recurrent} as those are also used by the SOTA results on this task. Using layer-normalization and the recurrent dropout improves the performance significantly and reduces the effects of overfitting. We train our models with Adam \citep{kingma2014adam} over the sequences of length 150. We show our results in Table \ref{tbl:ptb_results}.

In addition to the regular char-LM experiments, in order to confirm our hypothesis regarding to the ability of MANNs generalizing to the sequences longer than the ones seen during the training. We have trained a language model which learns $p(\vy_t|\vh_t, \vr_t; \TT)$  by using a softmax layer as described in Equation \ref{eqn:prediction}. However to measure the performance of $p(\vy_t|\vr_t, \vx_t; \TT)$ on test set, we have used the softmax layer that gets into the auxiliary cost defined for the REINFORCE as in Equation \ref{eqn:auxilary_cost_reinforce} for a model trained with REINFORCE and with the auxiliary cost. As in Table \ref{tbl:ptb_results}, the model's performance by using $p(\vy_t|\vh_t, \vr_t; \TT)$ is 1.26, however by using $p(\vy_t|\vh_t, \vr_t; \TT)$ it becomes 1.28. This gap is small enough to confirm our assumption that $p(\vy_t|\vh_t, \vr_t; \TT) \approx p(\vy_t|\vr_t, \vx_t;\TT)$.


\begin{table}[htb]
    \centering
    \begin{tabular}{c c}
            \Xhline{0.8pt}
            \hline
            \bf Model & {\bf BPC} \\
            \hline
            CW-RNN~\citep{koutnik2014clockwork}                 & 1.46 \\
            HF-MRNN~\citep{sutskever2011generating}                  & 1.41 \\
            ME $n$-gram~\citep{mikolov2012subword}             & 1.37 \\
            BatchNorm LSTM~\citep{cooijmans2016recurrent}       & 1.32 \\
            Zoneout RNN~\citep{krueger2016zoneout}              & 1.27 \\
            LayerNorm LSTM~\citep{ha2016hypernetworks}           & 1.27 \\
            LayerNorm HyperNetworks~\citep{ha2016hypernetworks} & \textbf{1.23}\\
            LayerNorm HM-LSTM \& Step Fn. \& Slope Annealing\citep{chung2016hierarchical}     & 1.24 \\
            \hline
            Our LSTM + Layer Norm + Dropout & 1.28 \\
            TARDIS + REINFORCE + R & 1.28 \\
            TARDIS + REINFORCE + Auxiliary Cost & 1.28 \\
            TARDIS + REINFORCE + Auxiliary Cost + R & 1.26 \\
            TARDIS + Gumbel Softmax + ST + R& \textbf{1.25} \\
            \Xhline{0.8pt}
    \end{tabular}
    \caption{Character-level language modelling results on Penn TreeBank Dataset. TARDIS with Gumbel Softmax and straight-through~(ST) estimator performs better than REINFORCE and it performs competitively compared to the SOTA on this task. "+ R" notifies the use of RESET gates $\alpha$ and $\beta$.}
    \label{tbl:ptb_results}
\end{table}

\subsection{Sequential Stroke Multi-digit MNIST task}
\label{sec:mnist_strokes}

In this subsection, we introduce a new pen-stroke based sequential multi-digit MNIST prediction task as a benchmark for long term dependency modelling. We also benchmark the performance of LSTM and TARDIS in this challenging task.

\subsubsection{Task and Dataset}

Recently \citep{jong2016incre} introduced an MNIST pen stroke classification task and  also provided dataset which consisted of pen stroke sequences representing the skeleton of the digits in the MNIST dataset. Each MNIST digit image $\mathcal{I}$ is represented as a sequence of quadruples $\{dx_i,dy_i,eos_i,eod_i\}_{i=1}^{T}$, where $T$ is the number of pen strokes to define the digit, $(dx_i, dy_i)$ denotes the pen offset from the previous to the current stroke (can be 1, -1 or 0), $eos_i$ is a binary valued feature to denote end of stroke and $eod_i$ is another binary valued feature to denote end of the digit. In the original dataset, first quadruple contains absolute value $(x,y)$ instead of offsets $(dx,dy)$. Without loss of generality, we set the starting position $(x,y)$ to $(0,0)$ in our experiments. Each digit is represented by 40 strokes on an average and the task is to predict the digit at the end of the stroke sequence. 

While this dataset was proposed for incremental sequence learning in \citep{jong2016incre}, we consider the multi-digit version of this dataset to benchmark models that can handle long term dependencies. Specifically, given a sequence of pen-stroke sequences, the task is to predict the sequence of digits corresponding to each pen-stroke sequences in the given order. This is a challenging task since it requires the model to learn to predict the digit based on the pen-stroke sequence, count the number of digits and remember them and generate them in the same order after seeing all the strokes. In our experiments we consider 3 versions of this task with 5,10, and 15 digit sequences respectively. We generated 200,000 training data points by randomly sampling digits from the training set of the MNIST dataset. Similarly we generated 20,000 validation and test data points by randomly sampling digits from the validation set and test set of the MNIST dataset respectively. Average length of the stroke sequences in each of these tasks are 199, 399, and 599 respectively.

\begin{figure}[h]
\begin{center}
\includegraphics[scale=0.35]{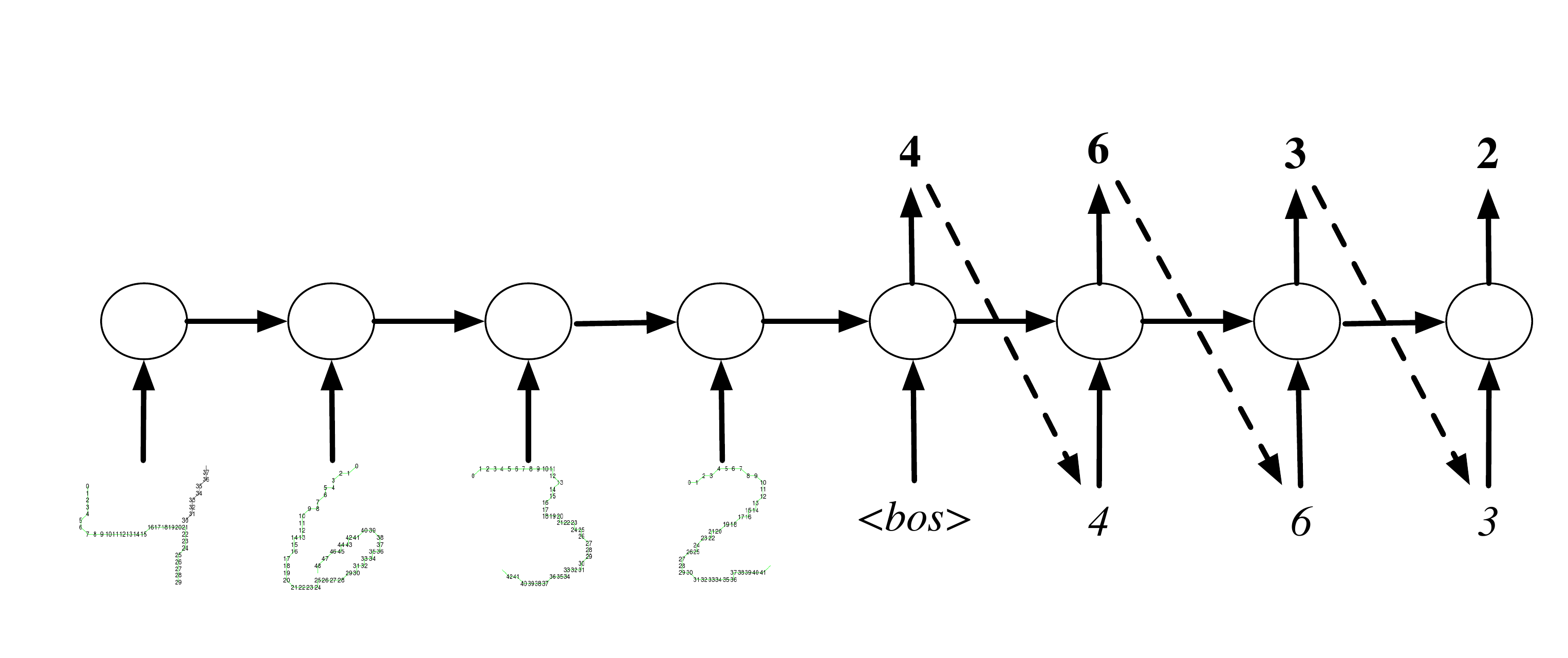}
\end{center}
\caption{An illustration of the sequential MNIST strokes task with multiple digits. The network is first provided with the sequence of strokes information for each MNIST digits(location information) as input, during the prediction the network tries to predict the MNIST digits that it has just seen. When the model tries to predict the predictions from the previous time steps are fed back into the network. For the first time step the model receives a special {\it <bos>} token which is fed into the model in the first time step when the prediction starts.}
\label{fig:mnist_strokes}
\end{figure}

\subsubsection{Results}

We benchmark the performance of LSTM and TARDIS in this new task. Both models receive the sequence of pen strokes and at the end of the sequence are expected to generate the sequence of digits followed by a particular {\it <bos>} token. The tasks is illustrated in Figure \ref{fig:mnist_strokes}. We evaluate the models based on per-digit error rate. We also compare the performance of TARDIS with REINFORCE with that of TARDIS with gumbel softmax. All the models were trained for same number of updates with early stopping based on the per-digit error rate in the validation set. Results for all 3 versions of the task are reported in Table-\ref{table:stroke}. From the table, we can see that TARDIS performs better than LSTM in all the three versions of the task. Also TARDIS with gumbel-softmax performs slightly better than TARDIS with REINFORCE, which is consistent with our other experiments.

\begin{table}[htbp]
\centering
\scalebox{1}{
\begin{tabular}{l|l|l|l}
\hline
\textbf{Model} & \textbf{5-digits} & \textbf{10-digits} & \textbf{15-digits}\\

\hline
\hline
LSTM & 3.00\% & 3.54\% & 8.81\% \\

TARDIS with REINFORCE & 2.09\% & 2.56\% & 3.67\% \\

TARDIS with gumbel softmax & \textbf{1.89\%} & \textbf{2.23\%} & \textbf{3.09\%} \\
\hline
\end{tabular}
}
\caption{Per-digit based test error in sequential stroke multi-digit MNIST task with 5,10, and 15 digits.}
\label{table:stroke}
\vskip -1em
\end{table}

\begin{figure}[h]      
\begin{center}
    \includegraphics[scale=0.32]{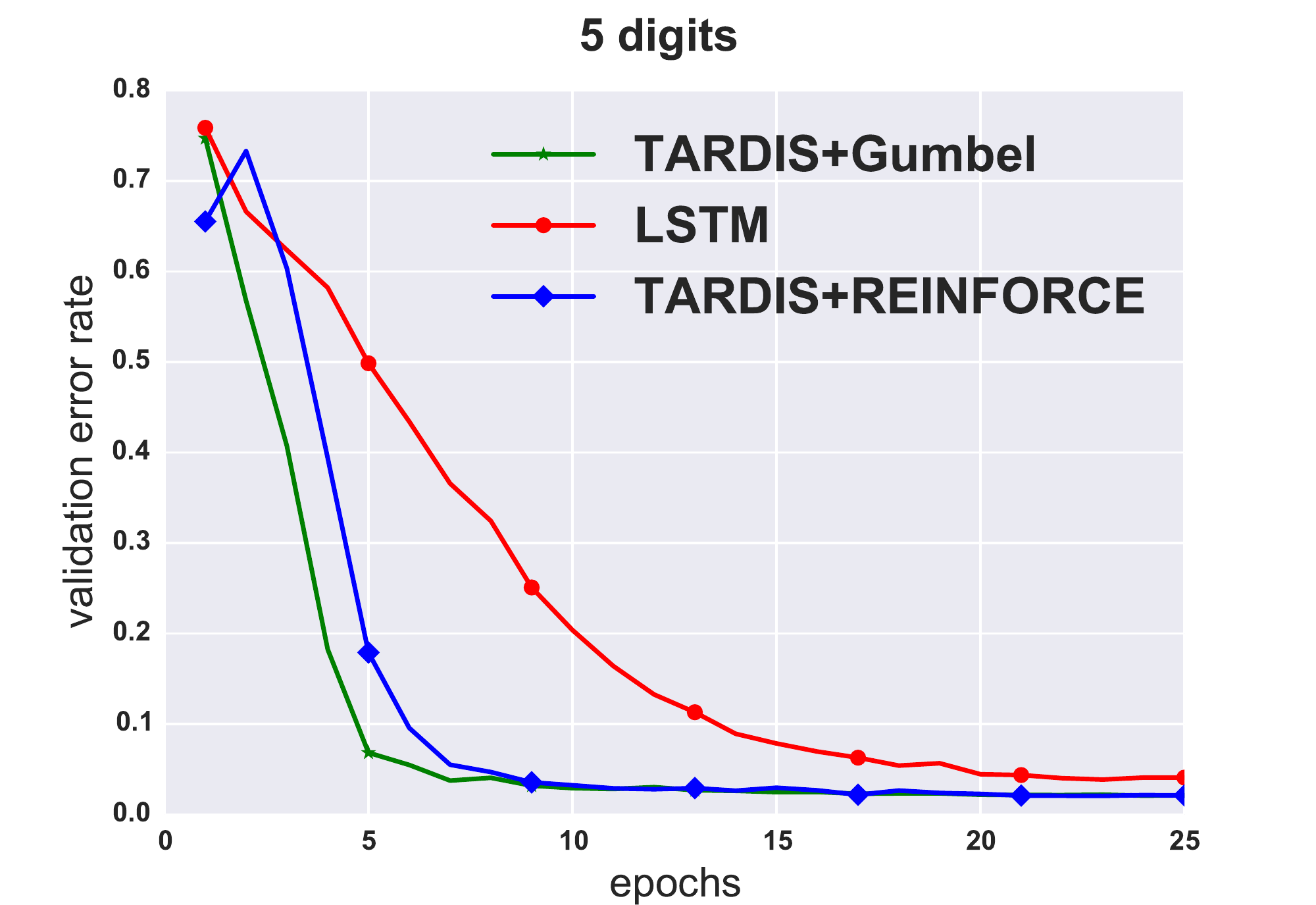} 
    \includegraphics[scale=0.32]{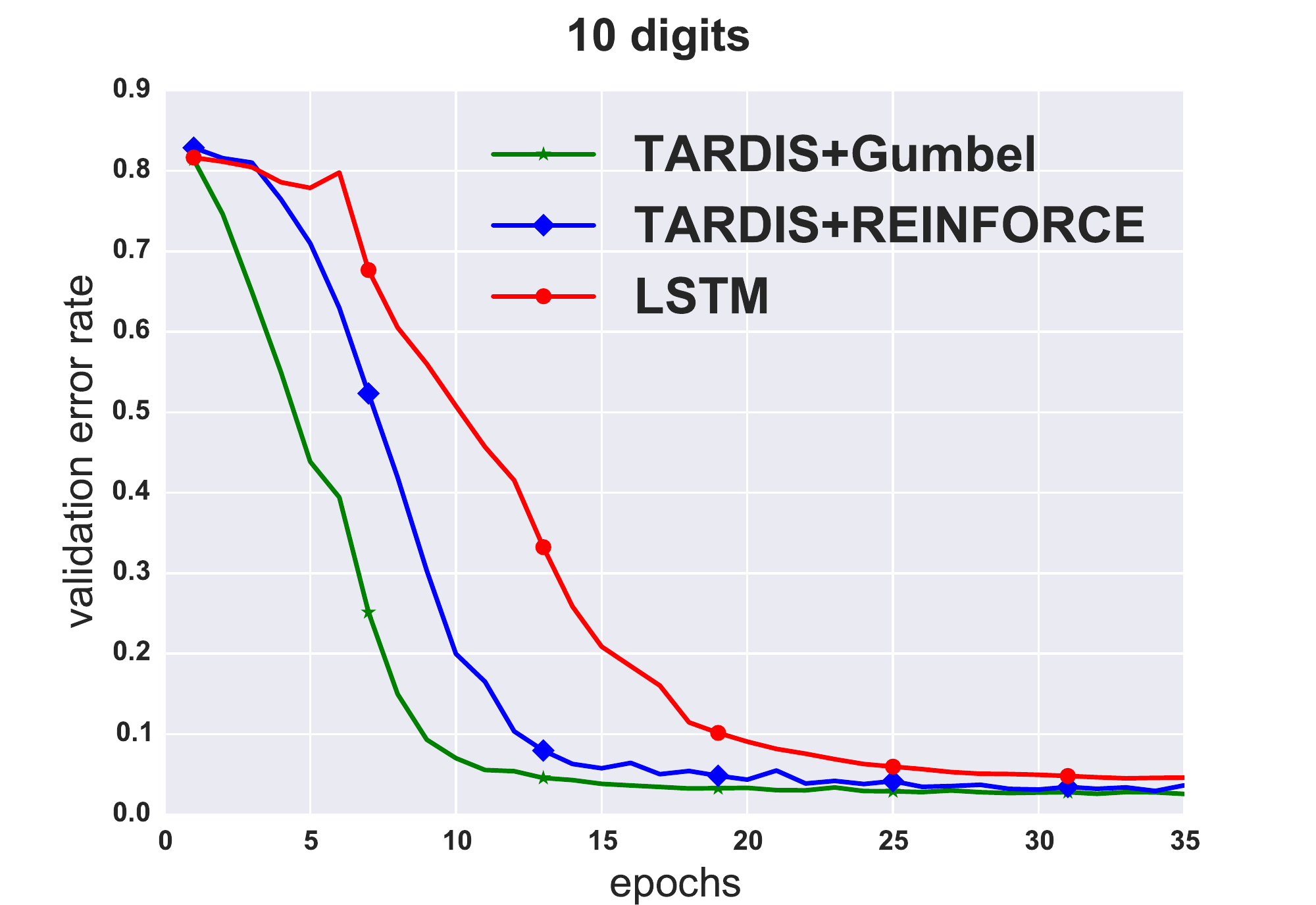} 
    \includegraphics[scale=0.32]{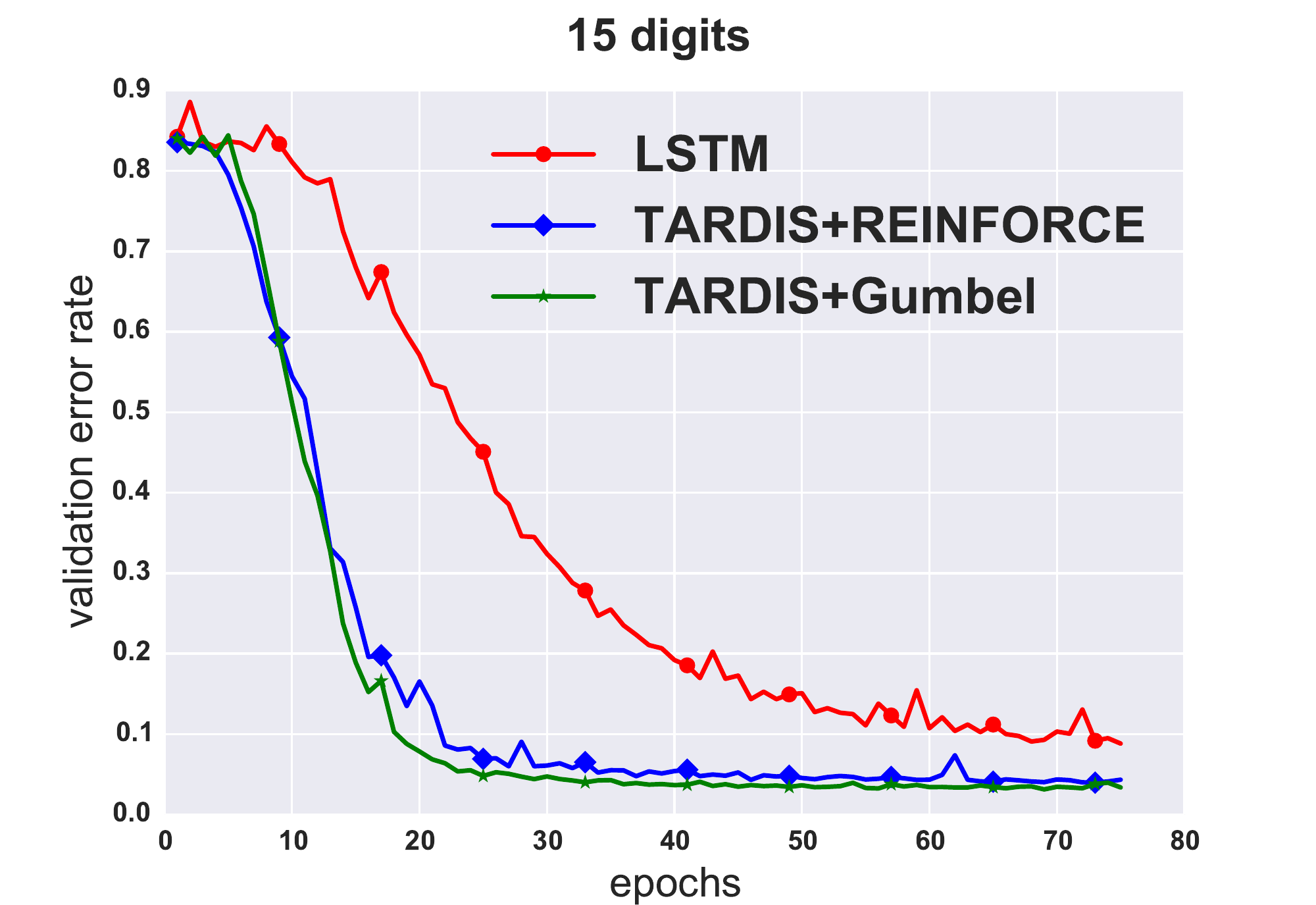} 
    \caption{Learning curves for LSTM and TARDIS for sequential stroke multi-digit MNIST task with 5, 10, and 15 digits respectively.}
    \label{sgraphs}
    \end{center}
\end{figure}

We also compare the learning curves of all the three models in Figure-\ref{sgraphs}. From the figure we can see that TARDIS learns to solve the task faster that LSTM by effectively utilizing the given memory slots. Also, TARDIS with gumbel softmax converges faster than TARDIS with REINFORCE.

\subsection{NTM Tasks}
\label{sec:ntm_tasks}

\cite{graves2014neural} proposed associative recall and the copy tasks to evaluate a model's ability to learn simple algorithms and generalize to the sequences longer than the ones seen during the training. We trained a TARDIS model with 4 features for the address and 32 features for the memory content part of the model. We used a model with hidden state of size 120. Our model uses a memory of size 16. We train our model with Adam and used the learning rate of 3e-3. We show the results of our model in Table \ref{tbl:ntm_toy_tasks}. TARDIS model was able to solve the both tasks, both with Gumbel-softmax and REINFORCE.

\begin{table}[htbp]
\centering
\footnotesize
\begin{tabular}{@{}lll@{}}
\toprule
           & Copy Task & Associative Recall \\ \midrule
D-NTM cont. \citep{gulcehre2016dynamic} & Success    & Success            \\
D-NTM discrete \citep{gulcehre2016dynamic} & Success    & Failure            \\
NTM   \citep{graves2014neural}     & Success     & Success            \\
\hline 
TARDIS + Gumbel Softmax + ST &  Success & Success \\
TARDIS  REINFORCE + Auxiliary Cost &  Success & Success \\
\bottomrule
\end{tabular}
\caption{In this table, we consider a model to be successful on copy or associative recall if its validation cost (binary cross-entropy) is lower than 0.02 over the sequences of maximum length seen during the training. We set the threshold to 0.02 to determine whether a model is successful on a task as in \citep{gulcehre2016dynamic}.}
\label{tbl:ntm_toy_tasks}
\end{table}


\subsection{Stanford Natural Language Inference}
\label{sec:snli_task}

\cite{bowman2015large} proposed a new task to test the machine learning algorithms' ability to infer whether two given sentences entail, contradict or are neutral(semantic independence) from each other. However, this task can be considered as a long-term dependency task, if the premise and the hypothesis are presented to the model in sequential order as also explored by \cite{rocktaschel2015reasoning}. Because the model should learn the dependency relationship between the hypothesis and the premise. Our model first reads the premise, then the hypothesis and at the end of the hypothesis the model predicts whether the premise and the hypothesis contradicts or entails. The model proposed by \cite{rocktaschel2015reasoning}, applies attention over its previous hidden states over premise when it reads the hypothesis. In that sense their model can still be considered to have some task-specific architectural design choice. TARDIS and our baseline LSTM models do not include any task-specific architectural design choices. In Table \ref{tbl:snli_results}, we compare the results of different models. Our model, performs significantly better than other models. However recently it has been shown that with architectural tweaks, it is possible to design a model specifically to solve this task and achieve 88.2\% test accuracy~\citep{chen2016enhancing}. 

\begin{table}[htb]
    \centering
    \begin{tabular}{c c}
            \Xhline{0.8pt}
            \hline
            \bf Model & {\bf Test Accuracy} \\
            \hline
            Word by Word Attention\citep{rocktaschel2015reasoning} & 83.5 \\
            Word by Word Attention two-way\citep{rocktaschel2015reasoning} & 83.2 \\
            \hline
            LSTM + LayerNorm + Dropout & 81.7 \\
            TARDIS + REINFORCE + Auxiliary Cost & 82.4 \\
            TARDIS + Gumbel Softmax + ST& \textbf{84.3} \\
            \Xhline{0.8pt}
    \end{tabular}
    \caption{Comparisons of different baselines on SNLI Task.}
    \label{tbl:snli_results}
\end{table}

\section{Conclusion}
\label{sec:conc}
In this paper, we propose a simple and efficient memory augmented neural network model which can perform well both on algorithmic tasks and more realistic tasks. Unlike the previous approaches, we show better performance on real-world NLP tasks, such as language modelling and SNLI. We have also proposed a new task to measure the performance of the models dealing with long-term dependencies. 

We provide a detailed analysis on the effects of using external memory for the gradients and justify the reason why MANNs generalize better on the sequences longer than the ones seen in the training set. We have also shown that the gradients will vanish at a much slower rate (if they vanish) when an external memory is being used. Our theoretical results should encourage further studies in the direction of developing better attention mechanisms that can create {\it wormhole} connections efficiently. 
\subsubsection*{Acknowledgments}
We thank Chinnadhurai Sankar for suggesting the phrase "wormhole connections" and proof-reading the paper. We would like to thank Dzmitry Bahdanau for the comments and feedback for the earlier version of this paper. We would like to also thank the developers of Theano~\footnote{\url{http://deeplearning.net/software/theano/}}, for
developing such a powerful tool for scientific computing
\cite{2016arXiv160502688short}. We acknowledge the support of the following
organizations for research funding and computing support: NSERC, Samsung,
Calcul Qu\'{e}bec, Compute Canada, the Canada Research Chairs and CIFAR. SC is supported by a FQRNT-PBEEE scholarship.

\bibliography{main}

\begin{thebibliography}{50}
\providecommand{\natexlab}[1]{#1}
\providecommand{\url}[1]{\texttt{#1}}
\expandafter\ifx\csname urlstyle\endcsname\relax
  \providecommand{\doi}[1]{doi: #1}\else
  \providecommand{\doi}{doi: \begingroup \urlstyle{rm}\Url}\fi

\bibitem[Ba et~al.(2016)Ba, Kiros, and Hinton]{ba2016layer}
Jimmy~Lei Ba, Jamie~Ryan Kiros, and Geoffrey~E Hinton.
\newblock Layer normalization.
\newblock \emph{arXiv preprint arXiv:1607.06450}, 2016.

\bibitem[Bahdanau et~al.(2015)Bahdanau, Cho, and Bengio]{bahdanau2014neural}
Dzmitry Bahdanau, Kyunghyun Cho, and Yoshua Bengio.
\newblock Neural machine translation by jointly learning to align and
  translate.
\newblock \emph{In Proceedings Of The International Conference on
  Representation Learning (ICLR 2015)}, 2015.

\bibitem[Bengio et~al.(1994)Bengio, Simard, and Frasconi]{bengio1994learning}
Yoshua Bengio, Patrice Simard, and Paolo Frasconi.
\newblock Learning long-term dependencies with gradient descent is difficult.
\newblock \emph{Neural Networks, IEEE Transactions on}, 5\penalty0
  (2):\penalty0 157--166, 1994.

\bibitem[Bengio et~al.(2013)Bengio, L{\'e}onard, and
  Courville]{bengio2013estimating}
Yoshua Bengio, Nicholas L{\'e}onard, and Aaron Courville.
\newblock Estimating or propagating gradients through stochastic neurons for
  conditional computation.
\newblock \emph{arXiv preprint arXiv:1308.3432}, 2013.

\bibitem[Bordes et~al.(2015)Bordes, Usunier, Chopra, and
  Weston]{bordes2015large}
Antoine Bordes, Nicolas Usunier, Sumit Chopra, and Jason Weston.
\newblock Large-scale simple question answering with memory networks.
\newblock \emph{arXiv preprint arXiv:1506.02075}, 2015.

\bibitem[Bowman et~al.(2015)Bowman, Angeli, Potts, and
  Manning]{bowman2015large}
Samuel~R Bowman, Gabor Angeli, Christopher Potts, and Christopher~D Manning.
\newblock A large annotated corpus for learning natural language inference.
\newblock \emph{arXiv preprint arXiv:1508.05326}, 2015.

\bibitem[Chandar et~al.(2016)Chandar, Ahn, Larochelle, Vincent, Tesauro, and
  Bengio]{chandar2016hierarchical}
Sarath Chandar, Sungjin Ahn, Hugo Larochelle, Pascal Vincent, Gerald Tesauro,
  and Yoshua Bengio.
\newblock Hierarchical memory networks.
\newblock \emph{arXiv preprint arXiv:1605.07427}, 2016.

\bibitem[Chen et~al.(2016)Chen, Zhu, Ling, Wei, and Jiang]{chen2016enhancing}
Qian Chen, Xiaodan Zhu, Zhenhua Ling, Si~Wei, and Hui Jiang.
\newblock Enhancing and combining sequential and tree lstm for natural language
  inference.
\newblock \emph{arXiv preprint arXiv:1609.06038}, 2016.

\bibitem[Cheng et~al.(2016)Cheng, Dong, and Lapata]{cheng2016long}
Jianpeng Cheng, Li~Dong, and Mirella Lapata.
\newblock Long short-term memory-networks for machine reading.
\newblock \emph{arXiv preprint arXiv:1601.06733}, 2016.

\bibitem[Cho et~al.(2014)Cho, van Merrienboer, Gulcehre, Bougares, Schwenk, and
  Bengio]{cho2014learning}
Kyunghyun Cho, Bart van Merrienboer, Caglar Gulcehre, Fethi Bougares, Holger
  Schwenk, and Yoshua Bengio.
\newblock Learning phrase representations using rnn encoder-decoder for
  statistical machine translation.
\newblock \emph{arXiv preprint arXiv:1406.1078}, 2014.

\bibitem[Chung et~al.(2016)Chung, Ahn, and Bengio]{chung2016hierarchical}
Junyoung Chung, Sungjin Ahn, and Yoshua Bengio.
\newblock Hierarchical multiscale recurrent neural networks.
\newblock \emph{arXiv preprint arXiv:1609.01704}, 2016.

\bibitem[Cooijmans et~al.(2016)Cooijmans, Ballas, Laurent, and
  Courville]{cooijmans2016recurrent}
Tim Cooijmans, Nicolas Ballas, C{\'e}sar Laurent, and Aaron Courville.
\newblock Recurrent batch normalization.
\newblock \emph{arXiv preprint arXiv:1603.09025}, 2016.

\bibitem[de~Jong(2016)]{jong2016incre}
Edwin~D. de~Jong.
\newblock Incremental sequence learning.
\newblock \emph{arXiv preprint arXiv:1611.03068}, 2016.

\bibitem[Grave et~al.(2016)Grave, Joulin, and Usunier]{grave2016improving}
Edouard Grave, Armand Joulin, and Nicolas Usunier.
\newblock Improving neural language models with a continuous cache.
\newblock \emph{arXiv preprint arXiv:1612.04426}, 2016.

\bibitem[Graves et~al.(2014)Graves, Wayne, and Danihelka]{graves2014neural}
Alex Graves, Greg Wayne, and Ivo Danihelka.
\newblock Neural turing machines.
\newblock \emph{arXiv preprint arXiv:1410.5401}, 2014.

\bibitem[Graves et~al.(2016)Graves, Wayne, Reynolds, Harley, Danihelka,
  Grabska-Barwi\'{n}ska, Colmenarejo, Grefenstette, Ramalho, Agapiou, Badia,
  Hermann, Zwols, Ostrovski, Cain, King, Summerfield, Blunsom, Kavukcuoglu, and
  Hassabis]{Graves_Nature2016}
Alex Graves, Greg Wayne, Malcolm Reynolds, Tim Harley, Ivo Danihelka, Agnieszka
  Grabska-Barwi\'{n}ska, Sergio~G. Colmenarejo, Edward Grefenstette, Tiago
  Ramalho, John Agapiou, Adri\`{a}~P. Badia, Karl~M. Hermann, Yori Zwols, Georg
  Ostrovski, Adam Cain, Helen King, Christopher Summerfield, Phil Blunsom,
  Koray Kavukcuoglu, and Demis Hassabis.
\newblock {Hybrid computing using a neural network with dynamic external
  memory}.
\newblock \emph{Nature}, advance online publication, October 2016.
\newblock ISSN 0028-0836.
\newblock \doi{10.1038/nature20101}.
\newblock URL \url{http://dx.doi.org/10.1038/nature20101}.

\bibitem[Grefenstette et~al.(2015)Grefenstette, Hermann, Suleyman, and
  Blunsom]{grefenstette2015learning}
Edward Grefenstette, Karl~Moritz Hermann, Mustafa Suleyman, and Phil Blunsom.
\newblock Learning to transduce with unbounded memory.
\newblock In \emph{Advances in Neural Information Processing Systems}, pages
  1819--1827, 2015.

\bibitem[Gulcehre et~al.(2016)Gulcehre, Chandar, Cho, and
  Bengio]{gulcehre2016dynamic}
Caglar Gulcehre, Sarath Chandar, Kyunghyun Cho, and Yoshua Bengio.
\newblock Dynamic neural turing machine with soft and hard addressing schemes.
\newblock \emph{arXiv preprint arXiv:1607.00036}, 2016.

\bibitem[Ha et~al.(2016)Ha, Dai, and Le]{ha2016hypernetworks}
David Ha, Andrew Dai, and Quoc~V Le.
\newblock Hypernetworks.
\newblock \emph{arXiv preprint arXiv:1609.09106}, 2016.

\bibitem[Hochreiter(1991)]{hochreiter1991untersuchungen}
Sepp Hochreiter.
\newblock Untersuchungen zu dynamischen neuronalen netzen.
\newblock \emph{Diploma, Technische Universit{\"a}t M{\"u}nchen}, page~91,
  1991.

\bibitem[Hochreiter and Schmidhuber(1997)]{lstm1997}
Sepp Hochreiter and J{\"{u}}rgen Schmidhuber.
\newblock Long short-term memory.
\newblock \emph{Neural Computation}, 9\penalty0 (8):\penalty0 1735--1780, 1997.

\bibitem[Jang et~al.(2016)Jang, Gu, and Poole]{jang2016categorical}
Eric Jang, Shixiang Gu, and Ben Poole.
\newblock Categorical reparameterization with gumbel-softmax.
\newblock \emph{arXiv preprint arXiv:1611.01144}, 2016.

\bibitem[Joulin and Mikolov(2015)]{joulin2015inferring}
Armand Joulin and Tomas Mikolov.
\newblock Inferring algorithmic patterns with stack-augmented recurrent nets.
\newblock In \emph{Advances in Neural Information Processing Systems}, pages
  190--198, 2015.

\bibitem[Kaiser and Sutskever(2015)]{kaiser2015neural}
{\L}ukasz Kaiser and Ilya Sutskever.
\newblock Neural gpus learn algorithms.
\newblock \emph{arXiv preprint arXiv:1511.08228}, 2015.

\bibitem[Kingma and Ba(2014)]{kingma2014adam}
Diederik Kingma and Jimmy Ba.
\newblock Adam: A method for stochastic optimization.
\newblock \emph{arXiv preprint arXiv:1412.6980}, 2014.

\bibitem[Koutnik et~al.(2014)Koutnik, Greff, Gomez, and
  Schmidhuber]{koutnik2014clockwork}
Jan Koutnik, Klaus Greff, Faustino Gomez, and Juergen Schmidhuber.
\newblock A clockwork rnn.
\newblock \emph{arXiv preprint arXiv:1402.3511}, 2014.

\bibitem[Krueger et~al.(2016)Krueger, Maharaj, Kram{\'a}r, Pezeshki, Ballas,
  Ke, Goyal, Bengio, Larochelle, Courville, et~al.]{krueger2016zoneout}
David Krueger, Tegan Maharaj, J{\'a}nos Kram{\'a}r, Mohammad Pezeshki, Nicolas
  Ballas, Nan~Rosemary Ke, Anirudh Goyal, Yoshua Bengio, Hugo Larochelle, Aaron
  Courville, et~al.
\newblock Zoneout: Regularizing rnns by randomly preserving hidden activations.
\newblock \emph{arXiv preprint arXiv:1606.01305}, 2016.

\bibitem[Kuhn and De~Mori(1990)]{kuhn1990cache}
Roland Kuhn and Renato De~Mori.
\newblock A cache-based natural language model for speech recognition.
\newblock \emph{IEEE transactions on pattern analysis and machine
  intelligence}, 12\penalty0 (6):\penalty0 570--583, 1990.

\bibitem[Loyka(2015)]{loyka2015singular}
Sergey Loyka.
\newblock On singular value inequalities for the sum of two matrices.
\newblock \emph{arXiv preprint arXiv:1507.06630}, 2015.

\bibitem[Maddison et~al.(2016)Maddison, Mnih, and Teh]{maddison2016concrete}
Chris~J Maddison, Andriy Mnih, and Yee~Whye Teh.
\newblock The concrete distribution: A continuous relaxation of discrete random
  variables.
\newblock \emph{arXiv preprint arXiv:1611.00712}, 2016.

\bibitem[Marcus et~al.(1993)Marcus, Marcinkiewicz, and
  Santorini]{marcus1993building}
Mitchell~P Marcus, Mary~Ann Marcinkiewicz, and Beatrice Santorini.
\newblock Building a large annotated corpus of english: The penn treebank.
\newblock \emph{Computational linguistics}, 19\penalty0 (2):\penalty0 313--330,
  1993.

\bibitem[Mikolov et~al.(2012)Mikolov, Sutskever, Deoras, Le, Kombrink, and
  Cernocky]{mikolov2012subword}
Tom{\'a}{\v{s}} Mikolov, Ilya Sutskever, Anoop Deoras, Hai-Son Le, Stefan
  Kombrink, and J~Cernocky.
\newblock Subword language modeling with neural networks.
\newblock \emph{preprint (http://www. fit. vutbr. cz/imikolov/rnnlm/char.
  pdf)}, 2012.

\bibitem[Mnih and Gregor(2014)]{mnih2014neural}
Andriy Mnih and Karol Gregor.
\newblock Neural variational inference and learning in belief networks.
\newblock \emph{arXiv preprint arXiv:1402.0030}, 2014.

\bibitem[Pascanu et~al.(2013{\natexlab{a}})Pascanu, Gulcehre, Cho, and
  Bengio]{pascanu2013construct}
Razvan Pascanu, Caglar Gulcehre, Kyunghyun Cho, and Yoshua Bengio.
\newblock How to construct deep recurrent neural networks.
\newblock \emph{arXiv preprint arXiv:1312.6026}, 2013{\natexlab{a}}.

\bibitem[Pascanu et~al.(2013{\natexlab{b}})Pascanu, Mikolov, and
  Bengio]{pascanu2013difficulty}
Razvan Pascanu, Tomas Mikolov, and Yoshua Bengio.
\newblock On the difficulty of training recurrent neural networks.
\newblock \emph{ICML (3)}, 28:\penalty0 1310--1318, 2013{\natexlab{b}}.

\bibitem[Rae et~al.(2016)Rae, Hunt, Harley, Danihelka, Senior, Wayne, Graves,
  and Lillicrap]{rae2016scaling}
Jack~W. Rae, Jonathan~J. Hunt, Tim Harley, Ivo Danihelka, Andrew~W. Senior,
  Greg Wayne, Alex Graves, and Timothy~P. Lillicrap.
\newblock Scaling memory-augmented neural networks with sparse reads and
  writes.
\newblock \emph{CoRR}, abs/1610.09027, 2016.

\bibitem[Rockt{\"a}schel et~al.(2015)Rockt{\"a}schel, Grefenstette, Hermann,
  Ko{\v{c}}isk{\`y}, and Blunsom]{rocktaschel2015reasoning}
Tim Rockt{\"a}schel, Edward Grefenstette, Karl~Moritz Hermann, Tom{\'a}{\v{s}}
  Ko{\v{c}}isk{\`y}, and Phil Blunsom.
\newblock Reasoning about entailment with neural attention.
\newblock \emph{arXiv preprint arXiv:1509.06664}, 2015.

\bibitem[Santoro et~al.(2016)Santoro, Bartunov, Botvinick, Wierstra, and
  Lillicrap]{santoro2016one}
Adam Santoro, Sergey Bartunov, Matthew Botvinick, Daan Wierstra, and Timothy
  Lillicrap.
\newblock One-shot learning with memory-augmented neural networks.
\newblock \emph{arXiv preprint arXiv:1605.06065}, 2016.

\bibitem[Semeniuta et~al.(2016)Semeniuta, Severyn, and
  Barth]{semeniuta2016recurrent}
Stanislau Semeniuta, Aliaksei Severyn, and Erhardt Barth.
\newblock Recurrent dropout without memory loss.
\newblock \emph{arXiv preprint arXiv:1603.05118}, 2016.

\bibitem[Serban et~al.(2016)Serban, Sordoni, Bengio, Courville, and
  Pineau]{serban2016building}
Iulian~V Serban, Alessandro Sordoni, Yoshua Bengio, Aaron Courville, and Joelle
  Pineau.
\newblock Building end-to-end dialogue systems using generative hierarchical
  neural network models.
\newblock In \emph{Proceedings of the 30th AAAI Conference on Artificial
  Intelligence (AAAI-16)}, 2016.

\bibitem[Sukhbaatar et~al.(2015)Sukhbaatar, Szlam, Weston, and
  Fergus]{sukhbaatarend}
Sainbayar Sukhbaatar, Arthur Szlam, Jason Weston, and Rob Fergus.
\newblock End-to-end memory networks.
\newblock \emph{arXiv preprint arXiv:1503.08895}, 2015.

\bibitem[Sutskever et~al.(2011)Sutskever, Martens, and
  Hinton]{sutskever2011generating}
Ilya Sutskever, James Martens, and Geoffrey~E Hinton.
\newblock Generating text with recurrent neural networks.
\newblock In \emph{Proceedings of the 28th International Conference on Machine
  Learning (ICML-11)}, pages 1017--1024, 2011.

\bibitem[{Theano Development Team}(2016)]{2016arXiv160502688short}
{Theano Development Team}.
\newblock {Theano: A {Python} framework for fast computation of mathematical
  expressions}.
\newblock \emph{arXiv e-prints}, abs/1605.02688, May 2016.
\newblock URL \url{http://arxiv.org/abs/1605.02688}.

\bibitem[Trischler et~al.(2016)Trischler, Ye, Yuan, and
  Suleman]{trischler2016natural}
Adam Trischler, Zheng Ye, Xingdi Yuan, and Kaheer Suleman.
\newblock Natural language comprehension with the epireader.
\newblock \emph{arXiv preprint arXiv:1606.02270}, 2016.

\bibitem[Tulving(2002)]{tulving2002chronesthesia}
Endel Tulving.
\newblock Chronesthesia: Conscious awareness of subjective time.
\newblock 2002.

\bibitem[Weston et~al.(2015)Weston, Chopra, and Bordes]{weston2014memory}
Jason Weston, Sumit Chopra, and Antoine Bordes.
\newblock Memory networks.
\newblock \emph{In Proceedings Of The International Conference on
  Representation Learning (ICLR 2015)}, 2015.
\newblock In Press.

\bibitem[Williams(1992)]{williams92}
Ronald~J. Williams.
\newblock Simple statistical gradient-following algorithms for connectionist
  reinforcement learning.
\newblock \emph{Machine Learning}, 8:\penalty0 229--256, 1992.

\bibitem[Xu et~al.(2015)Xu, Ba, Kiros, Courville, Salakhutdinov, Zemel, and
  Bengio]{xu2015show}
Kelvin Xu, Jimmy Ba, Ryan Kiros, Aaron Courville, Ruslan Salakhutdinov, Richard
  Zemel, and Yoshua Bengio.
\newblock Show, attend and tell: Neural image caption generation with visual
  attention.
\newblock \emph{In Proceedings Of The International Conference on
  Representation Learning (ICLR 2015)}, 2015.

\bibitem[Zaremba and Sutskever(2015)]{rlntm}
Wojciech Zaremba and Ilya Sutskever.
\newblock Reinforcement learning neural turing machines.
\newblock \emph{CoRR}, abs/1505.00521, 2015.

\bibitem[Zilly et~al.(2016)Zilly, Srivastava, Koutn{\'\i}k, and
  Schmidhuber]{zilly2016recurrent}
Julian~Georg Zilly, Rupesh~Kumar Srivastava, Jan Koutn{\'\i}k, and J{\"u}rgen
  Schmidhuber.
\newblock Recurrent highway networks.
\newblock \emph{arXiv preprint arXiv:1607.03474}, 2016.

\end{thebibliography}

\end{document}